\documentclass{article}

\usepackage[T1]{fontenc}
\usepackage{url}
\usepackage{booktabs}
\usepackage{amsfonts}
\usepackage{amsmath}
\usepackage{amssymb}
\newcommand{\ind}[1]{\mathbf{1}_{#1}}
\usepackage{microtype}
\usepackage{graphicx}
\usepackage{xcolor}
\usepackage{multirow}
\usepackage{enumitem}
\usepackage{caption}
\usepackage{subcaption}
\usepackage{natbib}
\usepackage{pifont}
\usepackage{float}
\usepackage{placeins}
\definecolor{perplexityteal}{HTML}{20808D}
\usepackage[colorlinks=true,
            linkcolor=perplexityteal,
            citecolor=perplexityteal,
            urlcolor=perplexityteal]{hyperref}

\usepackage[margin=1in]{geometry}
\usepackage{times}
\usepackage{array}
\usepackage[most]{tcolorbox}
\usepackage{listings}
\definecolor{codebg}{HTML}{FBFCFC}
\definecolor{codekw}{HTML}{20808D}
\definecolor{codecomment}{HTML}{8A8A8A}
\definecolor{codestring}{HTML}{B25000}
\newtcblisting{pycode}{
  listing only, enhanced, breakable,
  colback=codebg, colframe=black!22, boxrule=0.5pt, arc=1.5pt,
  left=6pt, right=6pt, top=4pt, bottom=4pt,
  listing options={
    language=Python, basicstyle=\ttfamily\small,
    keywordstyle=\color{codekw}\bfseries, commentstyle=\color{codecomment}\itshape,
    stringstyle=\color{codestring}, showstringspaces=false,
    columns=fullflexible, keepspaces=true,
  }
}

\title{\normalfont WANDR: A Benchmark for Wide and Deep Research}

\makeatletter
\def\@maketitle{%
  \newpage
  \begin{center}%
    \rule{\textwidth}{1.5pt}\par\vspace{0.75em}%
    {\LARGE \@title \par}%
    \vspace{0.75em}%
    \rule{\textwidth}{0.5pt}\par%
    \vskip 1.5em%
    {
      Vitaliy Polshkov\thanks{Correspondence to \{vitalique, jeremy\}@perplexity.ai. We thank Omar Khattab for helpful suggestions.} \quad
      Marcin Pitera \quad
      Jeremy Yang\textsuperscript{*}  \\[0.9em]
      Kirill Priemko \quad
      Maksim Gaiduk \quad
      Aleksandr Nikolenko \quad \\[0.9em]
      Denis Bykov \quad
      Clare Southern \quad
      Denis Yarats \quad
      Jerry Ma \\[1.5em]
      Perplexity \\
    }%
    \vskip 1.5em
    July 16, 2026
  \end{center}%
  \par
  \vskip 1.5em}
\makeatother

\newtcolorbox{taskbox}[1][]{
  colback=perplexityteal!4, colframe=perplexityteal!60, boxrule=0.6pt,
  arc=2pt, left=6pt, right=6pt, top=5pt, bottom=5pt, fonttitle=\bfseries, #1}

\begin{document}

\maketitle

\begin{abstract}
WANDR (Wide ANd Deep Research) is a benchmark of 500 realistic and challenging agentic data-collection tasks for knowledge work. Each task asks a system to discover a large collection of members satisfying specified criteria (the \emph{wide} axis) and establish specific facts about each member through orchestrated web searches (the \emph{deep} axis). We represent a task as a qualification key hierarchy such as \texttt{company ($n$) $\rightarrow$ employee ($m$) $\rightarrow$ url ($k$)}, meaning ``find $n$ qualifying companies, $m$ qualifying employees per company, and $k$ qualifying source pages per employee.'' This hierarchy defines a target of $n \times m \times k$ \emph{records}. Instead of comparing submissions against a static gold answer set, a task-specific judge re-fetches the cited pages and verifies each record's claims against the cited evidence, allowing the benchmark to cover current and changing facts. Record-level verdicts are aggregated into soft and hard precision, recall, and F1 at the task level. Required record counts range from dozens to thousands, shifting the central challenge from finding a single correct answer or writing a polished report to open-ended discovery at high volume without sacrificing per-record factual correctness. Tasks are derived from de-identified product-usage logs and produced by a semi-automated pipeline. Automated checks and empirical audits screen them, with human review where needed; the final set is curated to stress-test recurring failure modes of frontier agentic search systems. The benchmark remains far from saturated: at a high effort setting, the strongest system achieves only 0.363 soft F1 and 0.133 hard F1. The tasks and evaluation harness are available at \url{https://github.com/perplexityai/wandr}.
\end{abstract}

\section{Introduction} \label{sec:intro}
Research agents are increasingly asked to produce not only a single answer or free-form report, but also a structured collection of facts, each requiring in-depth search to establish. This task shape recurs across the professional knowledge work observed in our production data, including competitive-landscape mapping, deal due diligence, literature review, market and product comparisons, and talent sourcing. It has two orthogonal dimensions: wide and deep. On the wide axis, the agent usually must discover a large set of entities not supplied in advance; some tasks instead specify a closed roster and require complete coverage. On the deep axis, it must enrich each entity through multiple orchestrated web searches. A representative request asks the agent to find at least 70 CEO or CFO appointments at US-based companies, each first announced during March--April 2026, and provide an authoritative appointment source for every company--appointee pair. A wide-and-deep research system returns such a structured collection, with one independently verifiable record per claim. WANDR tasks require dozens to thousands of such records. The central challenge is to achieve the requested volume through broad search while keeping every record factually correct.

To capture the variety of these use cases while keeping every claim independently verifiable, we represent each WANDR task as a flexible \emph{qualification key hierarchy}: a tree of identifying keys and filtering criteria. The running example, \path{ceo_cfo_appointments}, has a primary hierarchy \texttt{company(70) $\rightarrow$ company\_appointee(1) $\rightarrow$ url(1)}. More generally, each level may require several children---$n$ companies, $m$ appointees per company, and $k$ sources per appointee---so the hierarchy defines a target of $n \times m \times k$ records. Each record, the atomic unit of evaluation, cites a live page and includes verbatim excerpts, making it independently checkable.

At this scale, current agentic search systems often fail in predictable ways. We observe several recurring modes in production agent runs, some of which are also documented in prior literature: \emph{volume collapse} (incomplete query decomposition and insufficient retrieval at scale, sometimes followed by hallucinated values) \citep{lan2025deepwidesearch, wong2025widesearch}; \emph{snippet reliance} (using a search-result snippet without validating that its source supports the submitted claim) \citep{wong2025widesearch}; \emph{non-systematic constraint application} (applying a filter to the first few entities but silently dropping it thereafter); \emph{cross-reference skips} (asserting a claim without corroborating it across the required sources); \emph{premature stopping} (giving up after an initial search yields insufficient information) \citep{wong2025widesearch}; \emph{missing enrichment} (finding an entity but never gathering its required facts); and \emph{context overflow} (long search trajectories exhausting the available context) \citep{lan2025deepwidesearch}. WANDR originated as an internal benchmark designed to distinguish systems by how well they avoid these failure modes. In particular, it compares a \emph{Search as Code} (SaC) system, in which agents compose programmable retrieval primitives \citep{perplexity2026sac}, with conventional systems that distribute fixed-endpoint search-and-read loops across parallel subagents. SaC, the strongest system evaluated, achieves only 0.363 soft F1 and 0.133 hard F1, making the benchmark challenging even for frontier research agents.

The released task set instantiates recurring patterns from production data as self-contained benchmark tasks. Every admitted task must meet three top-level criteria: \emph{substantial volume} (usually hundreds of records and sometimes thousands); \emph{feasibility} (the requested volume is attainable); and \emph{difficulty and discrimination} (simple internal baselines score low while stronger systems score meaningfully higher). Together, these choices yield five defining characteristics whose combination distinguishes WANDR:
\begin{itemize}[leftmargin=1.4em, itemsep=1pt, topsep=2pt]

\item \textbf{Task shapes and topics grounded in production usage.} The released tasks are derived from de-identified requests observed in a real research product, preserving the domains, constraints, and wide-then-deep structures of professional knowledge work (Section~\ref{sec:task}).
\item \textbf{A compositional tree structure that flexibly encodes task shapes and makes failures localizable.} A flat list and a fixed-schema table are both special cases of the qualification key hierarchy. Failures can be localized across the tree: a system may fail to discover core task entities or members, enrich intermediate entities or keys, identify relevant pages, extract adequate evidence from those pages, or disambiguate identities when keys collide (Section~\ref{sec:anatomy}).
\item \textbf{A semi-automated task-construction pipeline.} Automated authoring checks, an empirical feasibility audit, and a judge audit establish the conditions needed for a functional task. Optional human review improves task quality but is not required for the task package to run and grade submissions. With no gold answer set to annotate, human effort is quality review rather than exhaustive answer construction (Section~\ref{sec:task}).
\item \textbf{Reference-free, evidence-verified grading.} Every record is a citation-backed claim (URL plus verbatim excerpts). A task-specific judge re-fetches the cited page. The full verdict evaluates both the page and submitted excerpts, while the retrieval-only verdict asks only whether the page satisfies every substantive task requirement. These verdicts are then aggregated into precision (accuracy among submitted members), recall (quality-adjusted completion relative to the required member count), and F1, each in soft (partial credit) and hard (binary pass/fail) forms (Section~\ref{sec:grading}, Appendix~\ref{appendix:criteria}).
\item \textbf{A scalable substrate for reinforcement learning (RL).} The task pipeline can generate diverse training packages; per-record verdicts and hierarchical metrics provide dense partial-progress rewards; per-level required counts provide an explicit difficulty gradient for curriculum learning; and the streaming, cache-aware grader can amortize overlapping work across training batches (Appendix~\ref{appendix:rl-training}).
\end{itemize}

The rest of the paper is organized as follows. Section~\ref{sec:related-work} situates WANDR among single-answer, collection, and domain-specialized benchmark families and compares how they establish ground truth. Section~\ref{sec:anatomy} defines the task structure and its variations; Section~\ref{sec:task} describes the task-construction pipeline and reports summary statistics for the released set; Section~\ref{sec:grading} details how submissions are graded; Section~\ref{sec:eval} evaluates six production systems; and Section~\ref{sec:discussion} summarizes the findings and discusses their implications, limitations, and future directions.

\section{Related Work} \label{sec:related-work}

We organize prior work on agentic information-seeking benchmarks primarily by output shape---a compact answer or a collection of items---and by how that output is graded. Domain specialization is a cross-cutting distinction: both compact-answer and collection benchmarks may restrict their tasks to one expert field.

\paragraph{Closed-ended browsing and question answering} The first family asks an agent to solve a bounded browsing or question-answering problem and return a compact response. BrowseComp poses deliberately obscure single-answer questions \citep{wei2025browsecomp}; Humanity's Last Exam targets expert-level, closed-ended academic questions \citep{phan2025humanity}; FRAMES and GAIA combine retrieval with multi-hop reasoning \citep{krishna2024frames, mialon2023gaia}; AssistantBench includes realistic web-research requests but permits at most five answers \citep{yoran2024assistantbench}; and VeriWeb decomposes long-horizon web tasks into chains of individually verifiable subtasks \citep{liu2025veriweb}. These benchmarks stress depth or end-to-end task completion, but their compact, gold-answer-based outputs do not test repeated depth across a large collection of discovered entities. WANDR instead makes that width--depth composition the scored object: every member must be discovered, enriched, and independently evidenced.

\paragraph{Wide and broad information seeking} Closest to WANDR are benchmarks whose deliverable is a collection of items graded individually. WideSearch \citep{wong2025widesearch} also sources tasks from real user queries and grades populated tables cell by cell, but primarily tests broad collection with comparatively shallow per-item fields. DeepSearchQA \citep{gupta2026deepsearchqa} includes both single- and set-valued answers, graded against gold answer sets. WideSeekBench \citep{huang2026wideseek} uses a multi-stage generation and quality-control pipeline to produce table-completion tasks with varied target volumes and logical constraints. GISA \citep{zhu2026gisa} spans item-, set-, list-, and table-valued answers and maintains a live subset through periodically updated answers. These benchmarks expand the width axis, but generally do not require the same multi-stage enrichment and corroboration process to be repeated for every discovered member. Most similar to WANDR, DeepWideSearch \citep{lan2025deepwidesearch} explicitly combines depth and width, converting 220 tasks from existing deep- and wide-search datasets and grading against human-verified ground-truth tables. WANDR differs in task shape and grading: its requests require wide-then-deep search to be repeated within one hierarchy, rather than combining separately sourced deep and wide task patterns; its grading verifies claims against submitted sources rather than relying on a gold solution. The related \emph{Table-as-Search} framework \citep{lan2026tableassearch} casts wide-and-deep search as table completion, though as a solving method rather than a benchmark.

\paragraph{Domain-specialized information seeking} Orthogonal to output format, a parallel line of work restricts search and research benchmarks to a single expert domain, testing domain knowledge alongside retrieval: medicine \citep{chen2025medbrowsecomp}, finance \citep{zhu2025findeepresearch, bigeard2025finance}, law \citep{li2025legalagentbench}, and academic literature \citep{zhou2025academicbrowse}. AutoResearchBench is also literature-specific: its \emph{Deep Research} track identifies a target paper, while its \emph{Wide Research} track collects all papers satisfying given conditions \citep{xiong2026autoresearchbench}. These benchmarks raise the expertise bar and often constrain the sources or tools used within one field. WANDR is instead domain-general: it applies a common hierarchical representation and grading framework to varied task structures across talent, finance, health, legal and regulatory work, and other professional domains.

\paragraph{Grading paradigms} Most prior benchmarks establish ground truth in one of three ways: matching gold labels for single-answer question answering; applying task-specific rubric criteria with a large language model (LLM) judge to long-form reports, as in our deep-research benchmark DRACO \citep{zhong2026draco}; or matching a gold collection in wide-search benchmarks. WANDR is the breadth-oriented counterpart to DRACO. Other work makes the evaluator itself agentic: Agent-as-a-Judge \citep{zhuge2024agent} evaluates code-agent outputs and trajectories against hierarchical requirements, while Mind2Web 2 \citep{gou2025mind2web} assesses answer correctness and source attribution for citation-backed responses. WANDR brings source verification to collection scale: every record must carry its own citation (URL plus verbatim excerpts), and a task-specific judge re-fetches the cited page and verifies the claim against it. Because the grader checks submitted claims rather than enumerating every eligible answer, this approach avoids exhaustive gold-answer annotation. Required-volume targets, validated for feasibility during task construction, provide the denominator for recall without requiring an enumerated answer set. This makes tasks about current and changing facts admissible and allows task production to scale through a semi-automated pipeline rather than exhaustive expert annotation. DeepWideSearch's own limitations call for exactly this combination of ``automated data generation techniques'' and ``reference-free evaluation metrics'' \citep{lan2025deepwidesearch}.

Table~\ref{tab:benchmark-comparison} summarizes WANDR's position. Relative to existing collection benchmarks, WANDR adds evidence-backed records, reference-free verification, and a compositional task structure: qualification key hierarchies in which discovery and per-entity enrichment form flexible trees rather than a single flat table. Its task shapes originate in real professional usage, and tasks are admitted only if they are high-volume, feasible, and discriminative against internal baselines. The resulting challenge is to sustain both breadth and per-record accuracy across a large, structured collection---precisely the regime in which current agents struggle.

\begin{table}[H]
\centering
\footnotesize
\renewcommand{\arraystretch}{1.15}
\begin{tabular}{@{}lccccc@{}}
\toprule
\textbf{Benchmark} & \textbf{Real-world} & \textbf{Open-set} & \textbf{Collection-record} & \textbf{At-scale quality} & \textbf{Reference-free} \\
 & \textbf{tasks} & \textbf{discovery} & \textbf{grading} & \textbf{task generation} & \textbf{verification} \\
\midrule
\multicolumn{6}{@{}l}{\emph{Closed-ended browsing \& question answering (task-level outputs)}} \\
GAIA \citep{mialon2023gaia}                 & \ding{55} & \ding{55} & \ding{55} & \ding{55} & \ding{55} \\
BrowseComp \citep{wei2025browsecomp}        & \ding{55} & \ding{55} & \ding{55} & \ding{55} & \ding{55} \\
FRAMES \citep{krishna2024frames}            & \ding{55} & \ding{55} & \ding{55} & \ding{55} & \ding{55} \\
AssistantBench \citep{yoran2024assistantbench} & \ding{55} & \ding{55} & \ding{55} & \ding{55} & \ding{55} \\
VeriWeb \citep{liu2025veriweb}              & \ding{55} & \ding{55} & \ding{55} & \ding{55} & \ding{55} \\
\midrule
\multicolumn{6}{@{}l}{\emph{Wide and broad information seeking (collections)}} \\
WideSearch \citep{wong2025widesearch}       & \ding{51} & \ding{51} & \ding{51} & \ding{55} & \ding{55} \\
DeepWideSearch \citep{lan2025deepwidesearch} & \ding{55} & \ding{51} & \ding{51} & \ding{55} & \ding{55} \\
DeepSearchQA \citep{gupta2026deepsearchqa}  & \ding{55} & \ding{51} & \ding{51} & \ding{55} & \ding{55} \\
AutoResearchBench (Wide) \citep{xiong2026autoresearchbench} & \ding{55} & \ding{51} & \ding{51} & \ding{55} & \ding{55} \\
WideSeekBench \citep{huang2026wideseek}     & \ding{55} & \ding{51} & \ding{51} & \ding{51} & \ding{55} \\
GISA \citep{zhu2026gisa}                    & \ding{55} & \ding{51} & \ding{51} & \ding{55} & \ding{55} \\
\textbf{WANDR (ours)}                        & \ding{51} & \ding{51} & \ding{51} & \ding{51} & \ding{51} \\
\bottomrule
\end{tabular}
\caption{WANDR versus representative search and research benchmarks. Rows are grouped by output format; domain specialization is orthogonal, so AutoResearchBench's Wide track appears with collection benchmarks. \textbf{Real-world tasks}: task shapes derived from real user requests or professional workflows rather than hand-curated puzzles or conversions of other benchmarks. \textbf{Open-set discovery}: the entities to report are not given and must be found at high recall. \textbf{Collection-record grading}: collection items, rows, or cells are a primary scoring unit, rather than a compact task-level answer or holistic report. \textbf{At-scale quality task generation}: a scalable generation process couples task creation with automated quality controls rather than relying only on manual task-by-task authoring. \textbf{Reference-free verification}: grading checks each claim against its fetched source, with no pre-annotated gold answer set to match. For mixed-format benchmarks, a check indicates that at least one named track or task family has the property.}
\label{tab:benchmark-comparison}
\end{table}
\section{Task Structure} \label{sec:anatomy}

\subsection{Recurring Patterns in Production Requests}

Wide-and-deep research requests vary widely in subject matter, but in the production data we observe a small number of recurring shapes. In \emph{entity discovery}, the system must find many instances of an entity class, such as companies, products, people, or events. In \emph{entity enrichment}, the entities are already known and the system must fill in the same facts for each one. The most common shape combines the two: first discover the entities, then investigate each one in depth. A fourth pattern is \emph{multi-condition research}, in which every candidate must satisfy several conditions, often supported by different pages or source types.

These patterns also appear inside one another. A request may ask for companies, several products per company, several facts per product, and one or more sources for every fact. Other requests repeat a fixed set of facets for each entity, compare the same entities across time, or ask both sides of a relationship to document the connection. The surface form changes, but the intent structure is the same: for each qualifying item, find a required set of qualifying children and carry the process through to source-backed evidence.

\subsection{A Common Tree Representation}

WANDR represents this repeated structure flexibly as a tree. Each level names the kind of item being collected, and each edge means \emph{for each parent, find these children}. Every branch ends in one or more URLs, so each claim remains tied to the page that supports it. This representation handles a wide range of real-world requests without defining a new output format for every workflow. The tree's topology specifies which items to find, how they are related, and how many are required. Concretely, it defines the keys, the parent--child edges, the required count at each level, and any subtasks whose top keys reuse keys in the parent hierarchy.

\subsection{Canonical Form} \label{sec:rhq}

A basic WANDR hierarchy has the form
\[
\texttt{key}_1(r_1) \rightarrow \texttt{key}_2(r_2) \rightarrow \cdots
\rightarrow \texttt{key}_d(r_d) \rightarrow \texttt{url}(r_u).
\]
A \emph{key} is the field, or group of fields, that identifies an item at one level. The value in parentheses is the minimum number of distinct children required at that level. A count is therefore a coverage floor; a closed-set level instead supplies the complete allowed roster. The hierarchy always ends in \texttt{url}. The count is applied separately beneath each parent: the running example asks for one appointment under each of 70 companies, not 70 appointments distributed across fewer companies. A key may be composite when one field is not globally unambiguous; \texttt{company\_appointee\{company, appointee\}} distinguishes appointments of the same person---or same-named people---across different companies.

A complete root-to-leaf path defines one record slot \texttt{\{item, url, excerpts, answer\}}: \texttt{item} contains the identifying key fields, \texttt{url} supplies the source page, \texttt{excerpts} contains verbatim passages selected from that page, and \texttt{answer} contains flexible JSON content describing the claim. A record is the atomic unit of grading. For corpus-level scale statistics, a \emph{member} is the task's designated core unit of coverage; in the running example, each company is one member and requires two leaf records, one in each branch. Counts apply recursively. A parent reaches its structural target when it has the required number of distinct children, and each child must in turn contain its own required descendants.

\subsection{Running Example}

We use the same \path{ceo_cfo_appointments} task for illustration.

\begin{taskbox}[title={\texttt{ceo\_cfo\_appointments}}]
\footnotesize
\textbf{Primary task:} For at least 70 US-based companies, identify at least one CEO or CFO appointment first announced between March 1 and April 30, 2026, and provide an authoritative appointment page.\\[3pt]
\textbf{Primary record:} one company--appointee appointment backed by an authoritative announcement page.\\[3pt]
\textbf{Primary validity and requirements:} the company is US-based; the page identifies the company and appointee, establishes the CEO or CFO role and first public announcement within the target window, and comes from an authoritative issuer, filing, newswire, or directly attributed first-hand business-journalism surface.\\[5pt]
\textbf{Company-listing subtask:} For the same 70 or more companies, supply at least one URL on a recognized listing-authority surface showing that the company has a primary or secondary listing on a US national securities exchange (NYSE, NASDAQ, NYSE American, or NYSE Arca) or is a US-domiciled SEC Exchange Act reporting issuer.\\[3pt]
\textbf{Subtask record:} the same company backed by a listing-authority page.\\[3pt]
\textbf{Listing requirements:}\\
1.\ The page identifies the claimed company.\\
2.\ The page communicates recognized listing-authority authorship, including through its URL when applicable.\\
3.\ The page shows the required US exchange listing or US-domiciled Exchange Act reporting status.\\[3pt]
\textbf{Records required:} 140 (70 appointment records plus 70 listing records).
\end{taskbox}

The primary hierarchy is
\[
\texttt{company(70)} \rightarrow \texttt{company\_appointee\{company, appointee\}(1)} \rightarrow \texttt{url(1)}.
\]
A subtask, \path{company_listings}, reuses the company key and adds one listing-authority URL per company. Its top key matches the primary root, so the two branches align at the company level and together require 70 appointment paths plus 70 listing paths.

\begin{figure}[h]
    \centering
    \includegraphics[width=\linewidth]{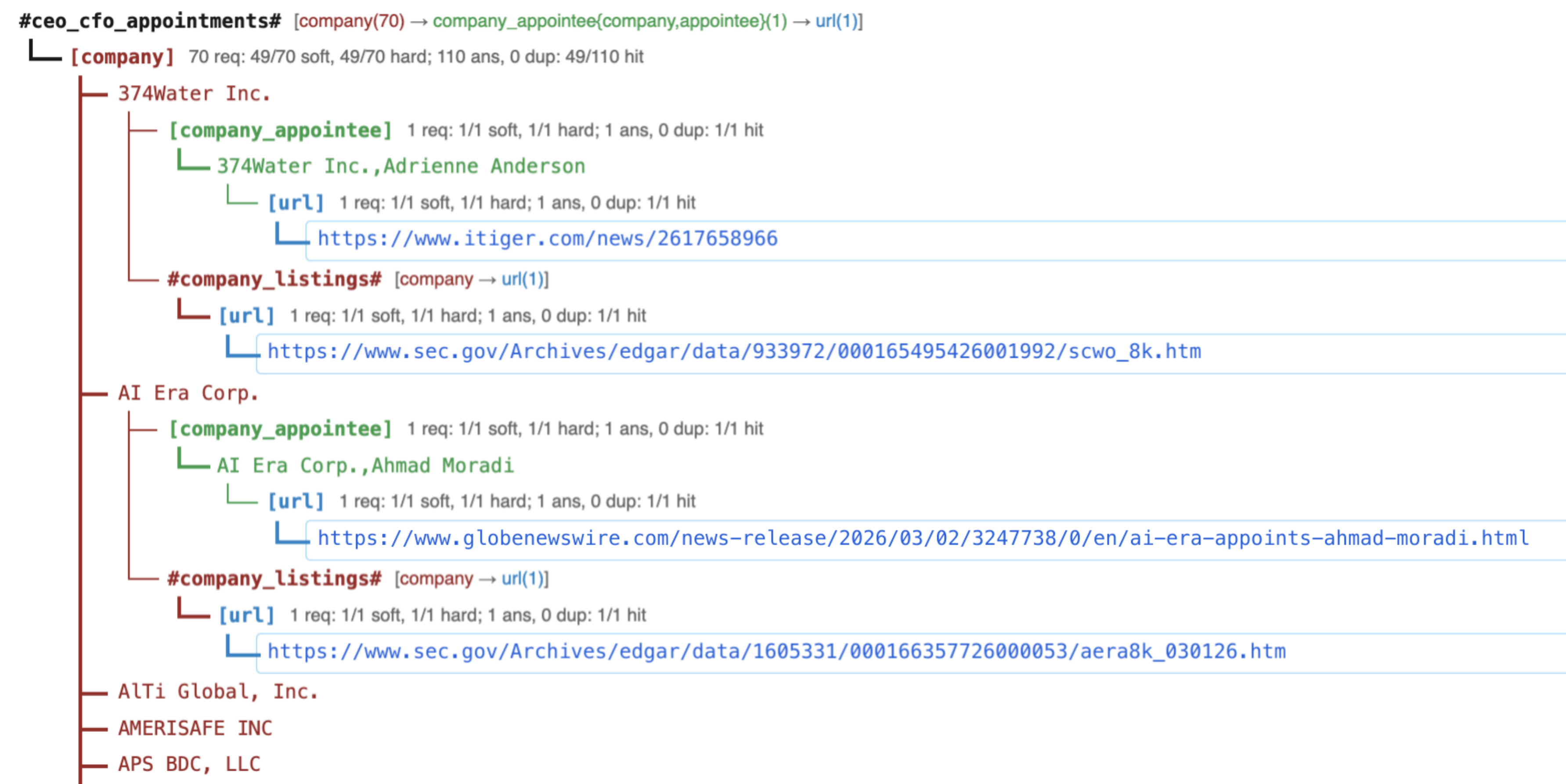}
    \caption{Task viewer for a graded \texttt{ceo\_cfo\_appointments} submission. The primary hierarchy requires 70 company--appointee records, and the \texttt{company\_listings} branch attaches one listing-authority URL to each shared company. The expanded rows show both branches for 374Water and AI Era; the remaining companies are collapsed.}
    \label{fig:ceo-cfo-tree}
\end{figure}
This one task demonstrates open-set discovery, a composite key, recursive per-parent counts, and a subtask. The next subsection uses shorter examples to explain other structural variations.

\subsection{Structural Variations} \label{sec:topo-patterns}
\begin{figure}[h]
    \centering
    \includegraphics[width=0.88\linewidth]{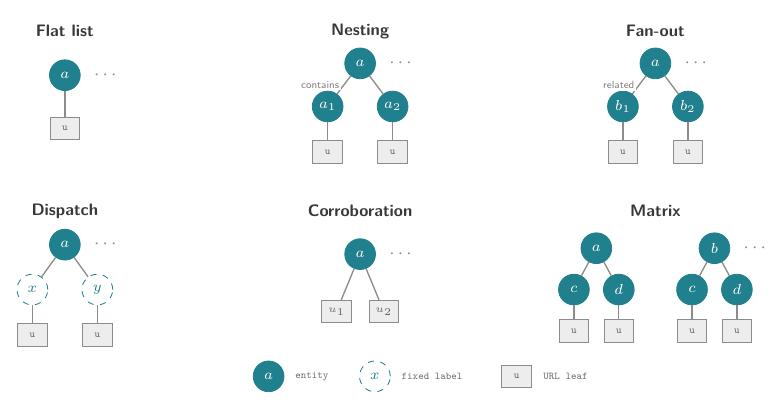}
    \caption{Six common hierarchy patterns. Filled circles are entities, dashed circles are fixed labels, and squares are URL leaves. In the matrix, repeated child labels denote the same child values under different parents.}
    \label{fig:topologies}
\end{figure}

The canonical hierarchy is intentionally small, but changing what a level represents, how many children it requires, or where a new branch attaches covers a broad range of research workflows. Figure~\ref{fig:topologies} highlights six common patterns; the four groups below define these patterns and related variations.

\paragraph{Open versus closed entity sets} Most levels ask the system to discover a requested number of qualifying values. A closed-set level instead supplies the complete roster, changing ``find $n$'' into ``find all.'' The hierarchy and scoring rules remain the same, but canonicalization and deduplication map submitted names to the supplied entities and enforce complete coverage; Appendix~\ref{appendix:identity} details these mechanisms.

\paragraph{Organizing entities}
\begin{itemize}[leftmargin=1.4em, itemsep=2pt, topsep=2pt]
    \item \emph{Flat list}: one entity level sits above the URL leaves, as in \texttt{company($n$) $\rightarrow$ url(1)}.
    \item \emph{Nesting}: levels follow a containment relationship, as in countries, cities within each country, and evidence for each city.
    \item \emph{Fan-out}: a parent connects to several related entities that are not contained within it. The \path{misquotation_instances} task asks for two attributed authors and five pages per quote--author pair; a startup may fan out to several investors.
    \item \emph{Matrix}: the same kind of child is repeated across many parents, often with an encouraged or fixed shared set. Examples include the same set of benchmarks for every model evaluated or the same set of products carried by every retailer.
\end{itemize}

\paragraph{Controlling coverage}
\begin{itemize}[leftmargin=1.4em, itemsep=2pt, topsep=2pt]
    \item \emph{Dispatch}: a level contains a fixed set of labels rather than discovered entities. The labels can represent facets, such as product, market, and customer review for a company. Requiring every label makes coverage explicit.
    \item \emph{Anchor}: a count-one intermediate key used to bind all downstream children to the same selected value and/or preserve the quota on its parent. In \texttt{company($n$) $\rightarrow$ product(1) $\rightarrow$ distributor(2)}, every company contributes a product and both distributors refer to that product; omitting the product level loses the shared-product guarantee, while flattening company and product loses per-company coverage.
    \item \emph{Partition}: a small top level divides the search space before the main entities are collected. \texttt{country(3) $\rightarrow$ company($m$)} enforces geographic spread that a flat company list would not.
\end{itemize}

\paragraph{Structuring evidence}
\begin{itemize}[leftmargin=1.4em, itemsep=2pt, topsep=2pt]
    \item \emph{Corroboration}: a URL count above one asks several distinct pages to support the same claim independently; each page must satisfy the full claim. A task may separately require source or domain independence.
    \item \emph{Triangulation}: several pages jointly describe a quantity or market surface without having to report the same value, such as three current retailer prices.
    \item \emph{Data as keys}: an extracted value becomes its own level when values must be counted, kept distinct, or held fixed across later evidence. A \texttt{price(1) $\rightarrow$ url(3)} branch requires all three pages to concern the same submitted price; a \texttt{signal($k$)} level requires $k$ distinct signals.
\end{itemize}

\paragraph{Adding subtasks} A subtask reuses a parent key but adds another hierarchy, with subtasks allowed to further recurse into subtasks of their own. They usually serve one of four roles:
\begin{itemize}[leftmargin=1.4em, itemsep=2pt, topsep=2pt]
    \item \emph{Enrichment}: add facts that are useful but separate from the primary claim, such as pricing for each model or a release paper for each benchmark.
    \item \emph{Another angle}: investigate a related question about the same entity, such as open roles alongside evidence of a recent company strategy change.
    \item \emph{Another source family}: require evidence from a distinct source universe, as the running CEO/CFO task does by separating appointment evidence from listing-authority evidence.
    \item \emph{Chain}: attach a later branch to entities identified by an earlier one, such as approval $\rightarrow$ active molecule $\rightarrow$ credited scientist $\rightarrow$ dissertation.
\end{itemize}

These variations are building blocks, not mutually exclusive task classes. A single hierarchy can combine nesting, fan-out, fixed dispatch labels, and multiple-source requirements, while subtasks add further branches at any shared key. This composability allows one representation to cover diverse workflows.

\section{Task Construction} \label{sec:task}
\begin{figure}[h]
    \centering
    \includegraphics[width=\linewidth]{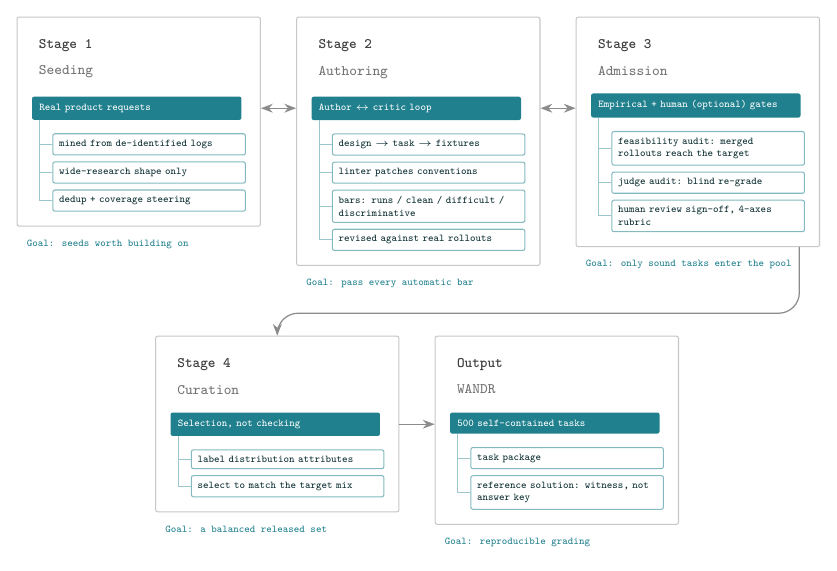}
    \caption{The task-construction pipeline has four stages. \textbf{Seeding} mines product logs for wide-research-shaped requests and selects a seed. \textbf{Authoring} uses an interleaved author--critic loop for design, task, and fixture drafts, plus a mechanical linter. \textbf{Admission} applies a feasibility audit based on merged rollouts and a judge audit, followed by optional human sign-off against the four-axis rubric; automated runnability, cleanliness, difficulty, and discrimination gates are enforced during authoring. \textbf{Curation} labels admitted tasks and selects a subset matching the target distributions. The output is 500 public task packages; system grading is a separate process (Section~\ref{sec:grading}).}
    \label{fig:task-construction}
\end{figure}

Tasks are produced by a semi-automated pipeline with four stages: \emph{seeding}, iterative \emph{authoring}, \emph{admission} into a candidate pool, and \emph{curation} into the released set (Figure~\ref{fig:task-construction}).

\emph{Seeding} identifies requests with a wide-research shape in de-identified product-usage logs. During \emph{authoring}, an author agent drafts the design, task specification, and fixtures, and a separate \emph{critic} reviews every draft; failures return to the stage that owns them. \emph{Admission} checks that the requested volume is attainable, audits the judge, and optionally adds human sign-off. Finally, \emph{curation} labels admitted tasks and selects a subset matching the target distributions. Each released task ships as a self-contained \emph{package} with its solver-facing task, verifier, fixtures, and labels. During construction, outputs from the authoring rollouts are merged into a best-known solution that serves as an internal feasibility witness, not an answer key; grading never compares a submission against it. Appendix~\ref{appendix:package-format} summarizes the public package format, while Appendix~\ref{appendix:admission-details} expands the admission gates. The subsections below describe seeding (Section~\ref{sec:sourcing}), authoring (Section~\ref{sec:factory}), admission (Section~\ref{sec:admission}), and curation (Section~\ref{sec:curation}); Section~\ref{sec:stats} then summarizes the released set.

\subsection{Seeding} \label{sec:sourcing}
Each task begins with a \emph{seed}: a real request mined from de-identified product-usage logs.\footnote{Two privacy safeguards apply. First, seed mining follows the de-identification methodology of the DRACO benchmark \citep{zhong2026draco}: requests are sampled from a de-identified pool and reworded to remove identifying details. Second, authoring retains only the seed's reusable pattern and re-grounds the task in entities with a public web presence. A released task therefore reflects a recurring form of professional work, not the content of any individual query or private record.} A first filter retains requests shaped like wide research---asking to compile a list, table, or directory---and favors sessions with heavy tool use. A classifier then ranks candidates by whether their entities and sources are stable enough to build on, their answers can be checked, and discovery is genuinely difficult. Rejected seeds remain as negative examples so they do not resurface. The authoring stage claims seeds in batches under two controls: \emph{deduplication} against previously used seeds and tasks, and \emph{coverage steering} toward the target distributions, which favors diverse, economically useful domains and caps overrepresented or low-value ones.

\subsection{Authoring} \label{sec:factory}

Authoring turns a seed's reusable pattern into a fresh, self-contained task through an iterative loop between two agents: an \emph{author} drafts, a separate \emph{critic} reviews, and the cycle repeats until the draft passes the relevant checks or returns to an earlier stage. A structural problem can send the work back to seeding. When a draft is too easy or exposes a shortcut, the author uses the topology patterns in Section~\ref{sec:topo-patterns} and the criteria and identity mechanisms in Appendices~\ref{appendix:criteria} and~\ref{appendix:identity} to raise its difficulty and adversarial robustness.\footnote{For example, if a seed permits a solver to cite mirror pages---duplicate copies of one listing on different domains---for every submission, the task is tightened to require distinct source domains per entity and prohibit template-substituted prose.} The work passes through three steps in order:\footnote{The mechanical linter runs between steps when only mechanical work remains. It patches convention violations without changing the entity class, evidence bar, volume, or judge policy. A meaning-changing fix is returned to the Task or Design step instead.}
\begin{itemize}[leftmargin=1.4em, itemsep=1pt, topsep=2pt]
\item \emph{Design}: the author sketches a few candidate directions for the task, with substantial web search to establish that the needed entities and pages actually exist; the critic selects one and sharpens it.
\item \emph{Task}: the author writes the task package---the key hierarchy, the description the solver reads, the judge specification, and the canonicalization/deduplication settings.
\item \emph{Fixture}: the author writes a few test records with known intended verdicts, grounded in real pages, and runs them through the judge to confirm the grader behaves as designed.\footnote{Fixtures are unit tests for the task's judge; they ship with the task but never grade the systems under test. When a fixture the author trusts disagrees with the judge, the disagreement is triaged---usually the fixture is wrong and gets fixed, but a genuine judge defect sends the task back for repair.}
\end{itemize}
Throughout the loop, four automated bars require that the package runs, its solver-facing description, judge specification, and schema remain aligned, primitive internal baselines do not already solve it, and stronger internal rollouts meaningfully outperform weaker ones. These are the runnability, cleanliness, difficulty, and discrimination checks shown in Figure~\ref{fig:task-construction}.

The authoring context also contains a set of vetted guidance tasks. Appendix~\ref{appendix:paragons} profiles a subset of five, chosen to expose the rationale behind matrix, open-set, temporal-panel, legal-comparison, and reciprocal-evidence structures.

The pipeline can also create \emph{sibling variants} by holding most of the hierarchy fixed while changing one controlled choice: a volume, time window, eligibility rule, or evidence bar. For example, \path{audio_gear} and its \path{audio_gear_relaxed} sibling share the same counts---50 products, two sentiment branches, and three URLs per branch---while the relaxed sibling accepts shorter and less dedicated opinion evidence. Sibling variants expose practical tradeoffs, such as admitting a larger target pool while increasing the required volume.

\subsection{Admission} \label{sec:admission}
Admission begins after a task passes the automated runnability, cleanliness, difficulty, and discrimination gates during authoring. It adds a feasibility audit, a judge audit, and optional human review. A failed check returns the task to the stage that owns the problem.

\paragraph{Feasibility audit} The pipeline merges everything found across 10--12 authoring rollouts into a single best-known solution, which must recover a near-full requested volume. A large surplus indicates that the task asks for too little; a persistent shortfall indicates that it asks for too much. This internal merged output is a feasibility witness, not an answer key. It establishes that the requested volume is attainable, while grading still evaluates every submission against its own cited pages rather than against the witness.

\paragraph{Judge audit} A reviewer samples graded records and regrades each one blindly, forming an independent verdict before seeing the automated judgment. Disagreements are attributed to solver behavior, task design, or grading machinery. Any major defect blocks the task. After cases attributable to solver behavior are separated, an automated-judge error rate above approximately $10\%$ of the remaining sample also blocks it.

\paragraph{Optional human review}\label{sec:meta-rubric} The author--critic loop automates much of the review; the human rubric provides a final checklist when sign-off is used. A reviewer scores four pass/fail axes: whether the \emph{key hierarchy} has sensible keys, an attainable volume floor, and appropriate URL corroboration; whether the \emph{task description} is complete and unambiguous; whether the \emph{grader} correctly encodes the judge instructions, schema, validity gates, and requirement checks; and whether \emph{identity handling} has appropriate canonicalization and deduplication settings. A reviewed task must pass all four axes. Appendix~\ref{appendix:admission-details} gives the complete gate and review tables.

\subsection{Curation} \label{sec:curation}
Curation selects a balanced release from the admitted pool. Because the pipeline deliberately overproduces, passing every check does not guarantee inclusion. Each admitted task receives distribution labels, and selection matches the target distributions; a strong task may remain unused when its bucket is already full. Selected tasks are then stamped into self-contained release packages. Across admission and curation, optional human work is budgeted at about 20 minutes per released task---approximately 15 minutes for rubric sign-off and 5 minutes for labeling---plus a roughly five-minute feasibility review for tasks that cannot be settled from merged rollouts. This is roughly one sixth of WideSearch's reported 2.33-hour average for one human completion; that study additionally used two independent annotators per task \citep{wong2025widesearch}. Omitting optional human review allows task production to scale further.

\subsection{Task Summary Statistics} \label{sec:stats}
We summarize the released benchmark by its subject areas, breadth, depth, and hierarchy. In this section, we focus on the static structural properties of the benchmark; Section~\ref{appendix:empirical-task-statistics} complements this analysis by examining how these requirements manifest in practice across historical rollouts.

\paragraph{Verticals} The release spans 13 vertical labels (Table~\ref{tab:verticals}). Each task has one to four labels: 294 tasks (58.8\%) have one label, and 206 (41.2\%) span multiple verticals. The mean is 1.50 labels per task.

\begin{table}[H]
    \centering
    \small
    \begin{tabular}{@{}lrr@{}}
        \toprule
        \textbf{Vertical} & \textbf{Tasks} & \textbf{Share} \\
        \midrule
        General    & 254 & 50.8\% \\
        Legal      & 168 & 33.6\% \\
        E-commerce &  99 & 19.8\% \\
        Technology &  66 & 13.2\% \\
        Events     &  55 & 11.0\% \\
        Finance    &  37 &  7.4\% \\
        Health     &  19 &  3.8\% \\
        Wikis      &  16 &  3.2\% \\
        Academic   &  15 &  3.0\% \\
        People     &   8 &  1.6\% \\
        Social     &   7 &  1.4\% \\
        Community  &   5 &  1.0\% \\
        Patents    &   3 &  0.6\% \\
        \bottomrule
    \end{tabular}
    \caption{Multi-label vertical coverage over all 500 benchmark tasks. Counts and shares are per label and therefore sum to more than 500 tasks and 100\%.}
    \label{tab:verticals}
\end{table}

\begin{figure}[h]
    \centering
    \includegraphics[width=.9\linewidth]{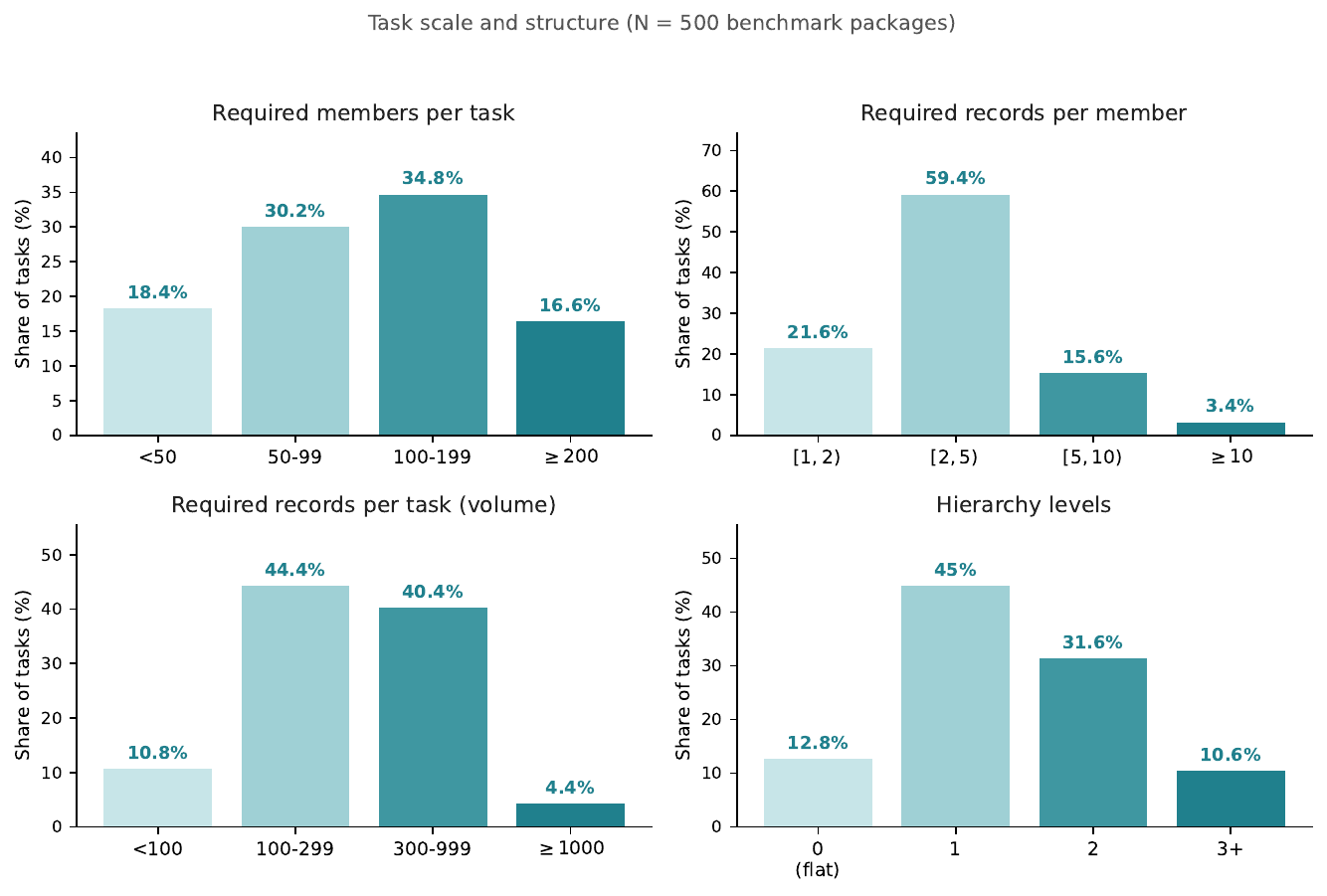}
    \caption{Required scale and structure of the 500-task benchmark set. Top left: required members under the designated task-level member keys. Top right: required records per member. Bottom left: required volume; for each task this is exactly the product of the two top-panel quantities. Bottom right: distinct non-member, non-URL keys across the root and all subtasks.}
    \label{fig:taskset-stats}
\end{figure}

\paragraph{Breadth vs.\ depth} For these corpus-level statistics, a member is the task's designated core unit of coverage, such as a company in the running example, whereas a record is one graded root-to-URL path. The required member count multiplies root quotas through the last designated member key, including intervening partition keys. Records per member is total required records divided by that count; their product therefore recovers required volume exactly (Figure~\ref{fig:taskset-stats}). The medians are 100 members, 3.00 records per member, and 245 records per task, respectively. Required volume is below 100 records for 10.8\% of tasks, 100--299 for 44.4\%, 300--999 for 40.4\%, and at least 1{,}000 for 4.4\%; across all tasks, the total is 170{,}495 records. The number of hierarchy levels counts distinct non-member, non-URL keys across the root and all subtasks: 12.8\% of tasks have none, 45.0\% have one, 31.6\% have two, and 10.6\% have three or more.

\begin{figure}[h]
    \centering
    \includegraphics[width=.82\linewidth]{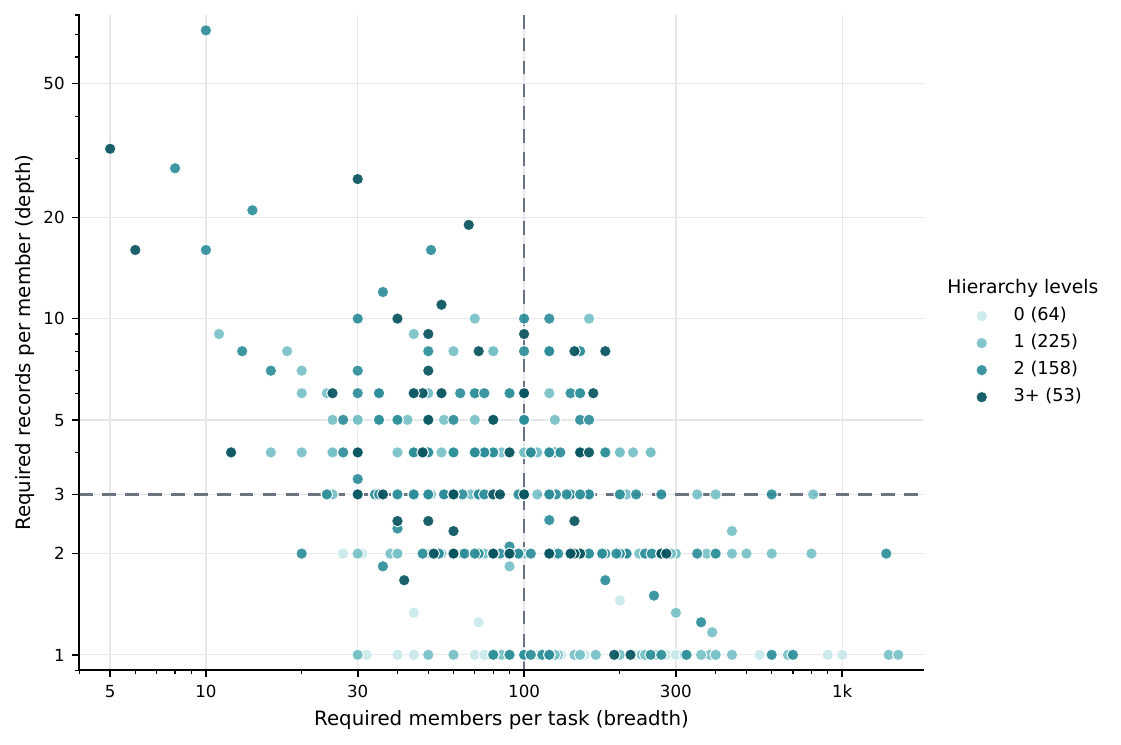}
    \caption{Task structure across required members and records per member (log--log; 500 benchmark tasks). Point color denotes the number of hierarchy levels: distinct non-member, non-URL keys across the root and all subtasks. Dashed guides mark the medians of 100 members and 3.00 records per member.}
    \label{fig:breadth-depth}
\end{figure}

\paragraph{Hierarchy and multiplicity} Figure~\ref{fig:breadth-depth} places the three structural quantities on one view. The number of hierarchy levels is not interchangeable with records per member: tasks with one, two, or at least three hierarchy levels all have a median of 3.00 records per member, while flat tasks have a median of 1.00. The former counts distinct key dimensions; the latter measures required record multiplicity after fixing member coverage.

Figures~\ref{fig:taskset-stats} and~\ref{fig:breadth-depth} report structural task statistics.

Structure alone does not capture source accessibility, semantic ambiguity, or the practical effort required to complete each branch. We additionally estimate complementary empirical statistics from the historical rollouts used during task construction; Appendix~\ref{appendix:empirical-task-statistics} is reserved for that analysis.

\section{Grading} \label{sec:grading}
Every leaf record receives an independent binary verdict, and the verdicts are then aggregated to the task level. The grader fetches the cited page and evaluates the submission \texttt{\{item, url, excerpts, answer\}} against universal and task-specific checks; a record passes only when every applicable condition holds. The judge prompt emits the repository field \path{verdict}; throughout the paper, we denote this complete-record verdict as \path{verdict_full}:
\begin{equation}\label{eq:verdict-full}
\begin{aligned}
\texttt{verdict\_full} = {} & \underbrace{\ind{\texttt{page\_content\_usable}} \times \ind{\texttt{answer\_intent\_clear}} \times \ind{\texttt{excerpts\_faithful}}}_{\text{universal}} \times {} \\
 & \underbrace{\scalebox{0.88}{$\displaystyle
 \ind{\texttt{overall\_valid}} \times \ind{\texttt{requirements\_all\_satisfied}} \times \ind{\texttt{requirements\_all\_supported}}
 $}}_{\text{task-specific}},
\end{aligned}
\end{equation}

The universal block applies the same evidence mechanics to every task:
\begin{itemize}[leftmargin=1.4em, itemsep=1pt, topsep=2pt]
\item \path{page_content_usable} requires substantive, on-topic page content.
\item \path{answer_intent_clear} requires a specific, identifiable submitted claim.
\item \path{excerpts_faithful} requires verbatim or near-verbatim, meaning-preserving excerpts.
\end{itemize}
The task-specific block combines eligibility and evidence:
\begin{itemize}[leftmargin=1.4em, itemsep=1pt, topsep=2pt]
\item \path{overall_valid} requires a well-formed, in-scope record.
\item \path{requirements_all_satisfied} asks whether the full page satisfies every substantive requirement.
\item \path{requirements_all_supported} asks whether the submitted excerpts alone support every requirement.
\end{itemize}

For comparison, the scorer defines a retrieval-only verdict:
\begin{equation}\label{eq:verdict-retrieval}
\texttt{verdict\_retrieval} = \ind{\texttt{requirements\_all\_satisfied}}.
\end{equation}
The full verdict, \path{verdict_full}, evaluates the complete submitted record using the full page and submitted excerpts. The retrieval-only verdict, \path{verdict_retrieval}, asks only whether the fetched page satisfies every substantive task requirement, independently of the submitted excerpts. Universal and validity fields are not mechanically multiplied into \path{verdict_retrieval}; instead, the judge is instructed to gate page-level satisfiability on those preceding checks when they are relevant. We use \path{verdict_full} for the main results and compare the two verdicts in Section~\ref{sec:failure-analysis}. Appendix~\ref{appendix:criteria} details the criterion schema.

The judge also reports confidence. A leaf signal is used only at confidence 2 or 3; lower-confidence leaves are treated as missing, so they do not enter precision but can reduce recall when their omission leaves fewer than the required number of members. Because grading re-fetches cited pages, solving and grading should occur close together: a URL that changes after submission can cause an otherwise correct record to fail.

\paragraph{Identity resolution} Before aggregation, the grader resolves identity independently along each key axis. Closed or predictable axes are canonicalized, while open-ended axes are semantically deduplicated; this prevents cosmetic variants from satisfying a volume floor more than once and makes coverage well-defined across parent tasks and subtasks. In the running example, the company key is resolved across the appointment and listing branches, while the composite company--appointee key distinguishes appointments. Appendix~\ref{appendix:identity} details the canonicalization, deduplication, and entity-scope policies.

\paragraph{Metrics} Leaf scores aggregate \emph{bottom-up}. At each level, precision averages every supplied child score. Recall first collapses duplicate identities by retaining the worst score for each entity, then sorts the distinct entity scores, keeps the top $k$, where $k$ is the required count, and zero-pads any shortfall before averaging. The grader reports soft and hard versions of precision, recall, and F1:\footnote{More detail is provided by the stylized rollup-algorithm implementation in Appendix~\ref{appendix:metric-rollup} and by a rollup-walkthrough example rendered via the repository-shipped viewer in Appendix~\ref{appendix:worked-scoring}.}
\begin{itemize}[leftmargin=1.4em, itemsep=1pt, topsep=2pt]
\item \textbf{Precision} --- the mean score among submitted members. Soft precision gives partial credit to incomplete members; hard precision assigns credit only when a submitted member's required subtree is fully correct.

\item \textbf{Recall} --- the sum of retained member scores divided by the required member count, measuring quality-adjusted completion relative to the target. If a submission exceeds the target, only the top $k$ members by score are retained; if it falls short, the remainder is zero-padded. Soft recall gives partial credit, while hard recall assigns credit only to submitted members whose required subtrees are fully correct.

\item \textbf{F1} --- the harmonic mean of precision and recall within each task. Soft F1 summarizes partial-credit collection quality; hard F1 summarizes complete-member performance.
\end{itemize}
The benchmark reports the simple unweighted mean of each per-task metric. Every task therefore contributes equally to the headline score, regardless of its required record count.

These metrics localize failure along several dimensions. \emph{Precision} measures the quality of submitted members, so low precision is consistent with a depth failure such as incorrect facts, unfaithful excerpts, or missing corroboration. The gap between scores under \path{verdict_retrieval} and \path{verdict_full} measures how much additional credit is lost when grading the complete submitted record rather than page-level task satisfaction alone. The drop from precision to recall is primarily a breadth signal: recall zero-pads unmet quotas and collapses duplicate identities, whereas precision averages the submitted children. A system that submits only a few perfect members can therefore earn high precision but low recall. Poor identity management can inflate raw submitted-member counts, but canonicalization and deduplication collapse aliases before recall is computed. The soft--hard gap is a completeness signal: hard scores remove partial credit from submitted members whose required subtrees are incomplete.

\section{Experiments and Results} \label{sec:eval}
\subsection{Setup}
\begin{figure}[h]
    \centering
    \includegraphics[width=.82\linewidth]{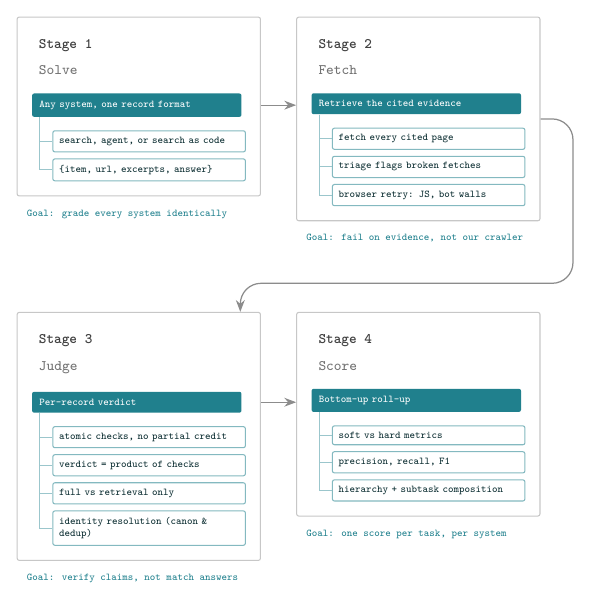}
    \caption{The grading pipeline. \textbf{Solve} normalizes each system's output into the common record format; \textbf{fetch} retrieves every cited page, with triage routing broken fetches to a browser retry; \textbf{judge} canonicalizes and deduplicates identities and emits a structured judgment for each record; \textbf{score} aggregates verdicts into soft and hard precision, recall, and F1 scores at the task level (Section~\ref{sec:grading}). The evaluated solver interfaces are described under Setup (Table~\ref{tab:systems}).}
    \label{fig:eval}
\end{figure}

Conceptually, evaluation uses a four-stage record-level pipeline---\emph{solve} $\rightarrow$ \emph{fetch} $\rightarrow$ \emph{judge} $\rightarrow$ \emph{score} (Figure~\ref{fig:eval}). \emph{Solve} converts every system's output into a common record format, ensuring identical downstream grading. \emph{Fetch} retrieves the cited pages and identifies broken or bot-walled results, including login walls, paywall stubs, and unrendered JavaScript. Triage routes these cases through a heavier JavaScript-rendering browser before judgment, reducing crawler-induced failures. In parallel, the evaluation pipeline canonicalizes and deduplicates each key axis. \emph{Judge} issues one verdict per record. \emph{Score} aggregates those verdicts into task-level metrics; during recall aggregation, duplicate identity variants collapse to one entity at their worst score. Append-only caching allows interrupted runs to resume rather than restart. Every released task packages the same pipeline.

We evaluate six production systems with different search and orchestration interfaces (Table~\ref{tab:systems}). These external systems are evaluated only after the task set is frozen and are not used for admission or rejection. The comparison includes task/search application programming interfaces (APIs; Exa and Parallel), general web-agent APIs (OpenAI's Responses API with web search and a code interpreter; Anthropic's managed agents), a deep-research agent (Gemini Deep Research), and a programmatic search-orchestration system (Perplexity Search as Code). All systems use the same pinned GPT-5.4 evaluation configuration: low effort for triage and canonicalization, medium for judging, and high for deduplication. Composite tasks multiply matching main-task and subtask scores, so a weak subtask reduces the composed entity score. By contrast, fan-out over a dispatch axis is averaged: covering three of four evidence types earns $0.75$, while a missing subtask cell can zero the corresponding entity.

\clearpage
\begin{table}[H]
\centering
\small
\begin{tabular}{@{}>{\raggedright\arraybackslash}p{0.13\linewidth}>{\raggedright\arraybackslash}p{0.19\linewidth}>{\raggedright\arraybackslash}p{0.25\linewidth}>{\raggedright\arraybackslash}p{0.13\linewidth}>{\raggedright\arraybackslash}p{0.17\linewidth}@{}}
\toprule
\textbf{System} & \textbf{Solver surface} & \textbf{Configuration} & \textbf{Model / setting} & \textbf{Output delivery} \\
\midrule
Perplexity  & Programmable primitives / orchestration (Search as Code) & Production Agent API & GPT-5.5 / high & File-sharing tool \\
Anthropic & General tool-agent endpoint & Managed Agents beta, 2026-04-01 toolset & Opus 4.8 / high & Sandbox file access \\
OpenAI & General tool-agent endpoint & Responses API, web search, code interpreter & GPT-5.5 / high & Sandbox file access \\
Gemini & Product deep-research interface & Gemini Deep Research, April 2026 preview & Deep Research / speed & Output tokens \\
Parallel & Hosted task/search endpoint & Parallel Tasks production API, processor \path{ultra4x} & --- / \path{ultra4x} & Output tokens \\
Exa & Hosted task/search endpoint & Exa Agent production API, effort \path{high} & --- / \path{high} & Output tokens \\
\bottomrule
\end{tabular}
\caption{Systems evaluated in the main benchmark run. Configurations, model settings, and output-delivery mechanisms are the recorded run settings; when the interface exposes no effort override, we report the selected model's documented default. Output delivery records how the harness obtains the large JSON submission: through provider sandbox files, Perplexity's file-sharing tool, or model output tokens. Solver-surface labels describe how the benchmark invokes each system; they are not product tiers or capability classes.}
\label{tab:systems}
\end{table}

\subsection{Main Results}

We report two sets of main results. On the full sample, we use the second-highest available setting for Perplexity, OpenAI, Parallel, and Exa, Gemini's \path{speed} setting,\footnote{We used Gemini \path{speed} instead of \path{max} on the full sample to save time and cost; \path{max} is evaluated in a 45-task subset.} and Anthropic's \path{high} setting,\footnote{Claude Managed Agents exposes no effort override in its agent configuration \citep{anthropic2026managedagents}, so we report effort from the selected model. Claude Opus 4.8 defaults the \path{effort} parameter to \path{high} across all surfaces, including the Messages API. See Anthropic's May 28, 2026 release notes: \url{https://platform.claude.com/docs/en/release-notes/overview\#may-28-2026}.} together with each system's best available delivery method. On a 45-task subset, we evaluate every available effort setting. Appendix~\ref{appendix:output-delivery} reports the delivery-method ablation on the same subset.

\paragraph{Full sample} We report one run per system over all 500 benchmark tasks. Scores are simple, unweighted means of per-task metrics. Trials ending in terminal errors after repeated retries have no metric-bearing verifier result and are zero-filled in the aggregate.\footnote{Retries target specific failure modes: provider/API retries handle request failures, Relay permits up to two full solver restarts, and verifier retries address incomplete judgments. A trial still errors if every attempt ends in a terminal provider failure, timeout, or missing required output.}

\begin{table}[H]
\centering
\small
\begin{tabular}{|cc|ccc|ccc|}
\hline
\multirow{2}{*}{\textbf{System}} & \multirow{2}{*}{\textbf{Completed}} & \multicolumn{3}{c|}{\textbf{Soft}} & \multicolumn{3}{c|}{\textbf{Hard}} \\
& & \textbf{Precision} & \textbf{Recall} & \textbf{F1} & \textbf{Precision} & \textbf{Recall} & \textbf{F1} \\
\hline
Perplexity & \textbf{500} & \textbf{0.389} & \textbf{0.357} & \textbf{0.363} & \textbf{0.150} & \textbf{0.134} & \textbf{0.133} \\
Anthropic  & \textbf{500} & \underline{0.354} & \underline{0.222} & \underline{0.249} & \underline{0.137} & \underline{0.073} & \underline{0.072} \\
OpenAI     & \underline{499} & 0.149 & 0.115 & 0.121 & 0.049 & 0.036 & 0.035 \\
Gemini     & 498 & 0.132 & 0.042 & 0.055 & 0.062 & 0.007 & 0.009 \\
Parallel   & 496 & 0.154 & 0.055 & 0.069 & 0.054 & 0.016 & 0.019 \\
Exa        & \textbf{500} & 0.130 & 0.057 & 0.070 & 0.042 & 0.018 & 0.020 \\
\hline
\end{tabular}
\caption{WANDR scores under \texttt{verdict\_full} for the main benchmark run. \textbf{Completed} is the scheduled 500 tasks minus errored trials and therefore equals the number of metric-bearing trials. Soft and hard precision, recall, and F1 are copied directly from the recorded job-level metrics; no score is reconstructed from another metric. Bold marks the best value in each numeric column and underline marks the second-best distinct value.}
\label{tab:results-scores}
\end{table}

\begin{table}[H]
\centering
\small
\begin{tabular}{|cc|c|cc|cc|}
\hline
\multirow{2}{*}{\textbf{System}} & \multirow{2}{*}{\textbf{Completed}} & \multicolumn{1}{c|}{\textbf{Cost}} & \multicolumn{2}{c|}{\textbf{Latency}} & \multicolumn{2}{c|}{\textbf{Token usage}} \\
& & \textbf{\$/task} & \textbf{Med. min} & \textbf{P90 min} & \textbf{In M} & \textbf{Out k} \\
\hline
Perplexity & \textbf{500} &  5.20 & 14.9 &  37.1 &  3.79 &  \textbf{29.6} \\
Anthropic  & \textbf{500} & 46.43 & 73.7 & 147.6 & 51.75 & 316.3 \\
OpenAI     & \underline{499} &  \textbf{0.50} &  \underline{8.6} &  \underline{12.9} &  \textbf{0.13} &  \underline{33.7} \\
Gemini     & 498 & 15.24 & 22.3 &  62.9 &  \underline{2.96} &  91.8 \\
Parallel   & 496 & 1.19 & 30.0 &  71.1 &    -- &    -- \\
Exa        & \textbf{500} &  \underline{0.50} & \textbf{5.3} &   \textbf{9.9} &    -- &    -- \\
\hline
\end{tabular}
\caption{Operational statistics for the same benchmark runs as Table~\ref{tab:results-scores}. \textbf{Completed} is the scheduled 500 tasks minus errored trials. \$/task is average solver cost. Med. min and P90 min are the median and 90th-percentile solve-stage latencies computed from first-to-last solve-stage timestamps, excluding verifier time. In M and Out k are average provider-reported input and output tokens per scheduled task, in millions and thousands; Exa and Parallel do not expose token counters. Bold marks the best value and underline marks the second-best distinct value, using unrounded values; higher is better for completion and lower is better for resource use.}
\label{tab:results-operational}
\end{table}

\begin{figure}[h]
    \centering
    \begin{subfigure}[t]{\linewidth}
        \centering
        \includegraphics[width=\linewidth]{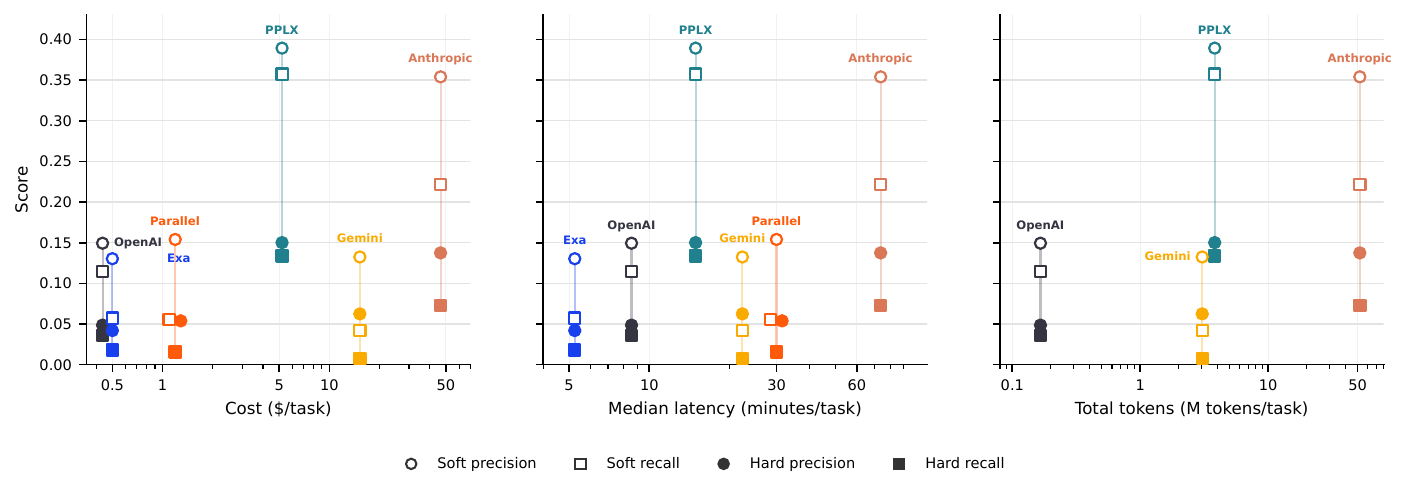}
        \caption{Soft and hard precision and recall.}
        \label{fig:performance-cost}
    \end{subfigure}
    \vspace{0.4em}
    \begin{subfigure}[t]{\linewidth}
        \centering
        \includegraphics[width=\linewidth]{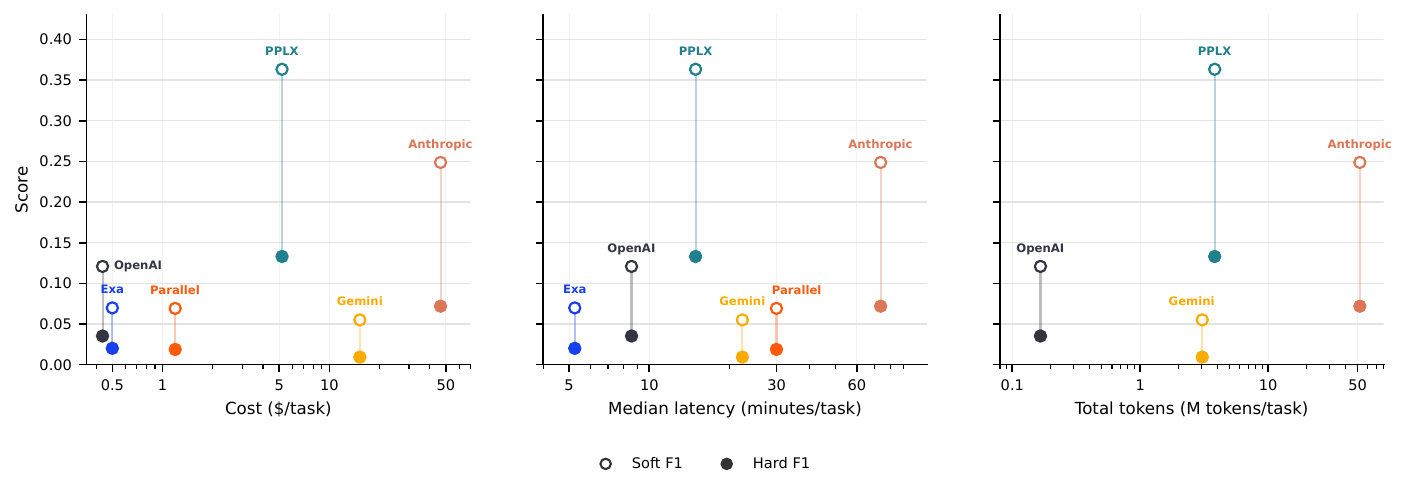}
        \caption{Soft and hard F1.}
        \label{fig:performance-cost-f1}
    \end{subfigure}
    \caption{Performance-cost views under \texttt{verdict\_full} for the main benchmark run. Each column uses a conventional log x-axis, so movement up and left is better. In (a), circles and squares denote precision and recall, while white and solid fill denote soft and hard metrics; in (b), fill denotes soft and hard F1. Small horizontal offsets are visual only; tables and prose report the unshifted values. Scores are the task-level means in Table~\ref{tab:results-scores}. Token use is the reported total per task; Exa and Parallel are omitted because they expose no token counters.}
    \label{fig:performance-cost-combined}
\end{figure}

Taken together, Tables~\ref{tab:results-scores} and~\ref{tab:results-operational} and Figure~\ref{fig:performance-cost-combined} show that no system dominates both performance and resource use. Perplexity leads at $0.363$ soft F1 and $0.133$ hard F1 with midrange cost (\$5.20 per task, $14.9$-minute median latency, and $3.82$M total reported tokens per task). Anthropic ranks second ($0.249$, $0.072$) but has the highest cost, latency, and token use. OpenAI and Exa cost about \$0.50 per task, and Exa is fastest at a $5.3$-minute median, but the remaining systems reach at most $0.121$ soft F1 and $0.035$ hard F1. Exa and Parallel do not expose token counts. The benchmark remains unsaturated: the highest mean task-level hard precision and recall are only $0.150$ and $0.134$, corresponding to complete credit for roughly one in seven submitted members and one in seven required members, respectively, under the benchmark's task-level averaging.

\paragraph{45-task subset} We evaluate every system--setting pair under the systems' best delivery channels, as in Table~\ref{tab:systems}. Tables~\ref{tab:supplemental-sweep-scores} and~\ref{tab:supplemental-sweep-operational} report the results. Perplexity, OpenAI, Parallel, and Exa each have four settings; Gemini has two and Anthropic one \path{high} configuration. \emph{Completed} is the number of metric-bearing trials out of 45.

\begin{figure}[h]
    \centering
    \begin{subfigure}[t]{\linewidth}
        \centering
        \includegraphics[width=\linewidth]{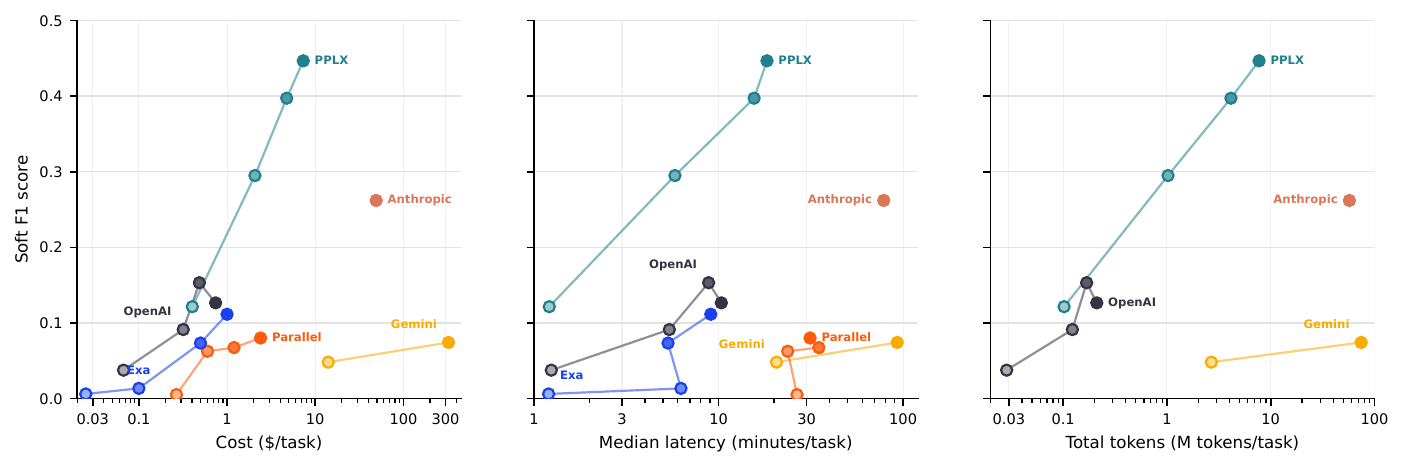}
        \caption{Soft F1.}
        \label{fig:sample-effort-soft-f1}
    \end{subfigure}
    \vspace{0.4em}
    \begin{subfigure}[t]{\linewidth}
        \centering
        \includegraphics[width=\linewidth]{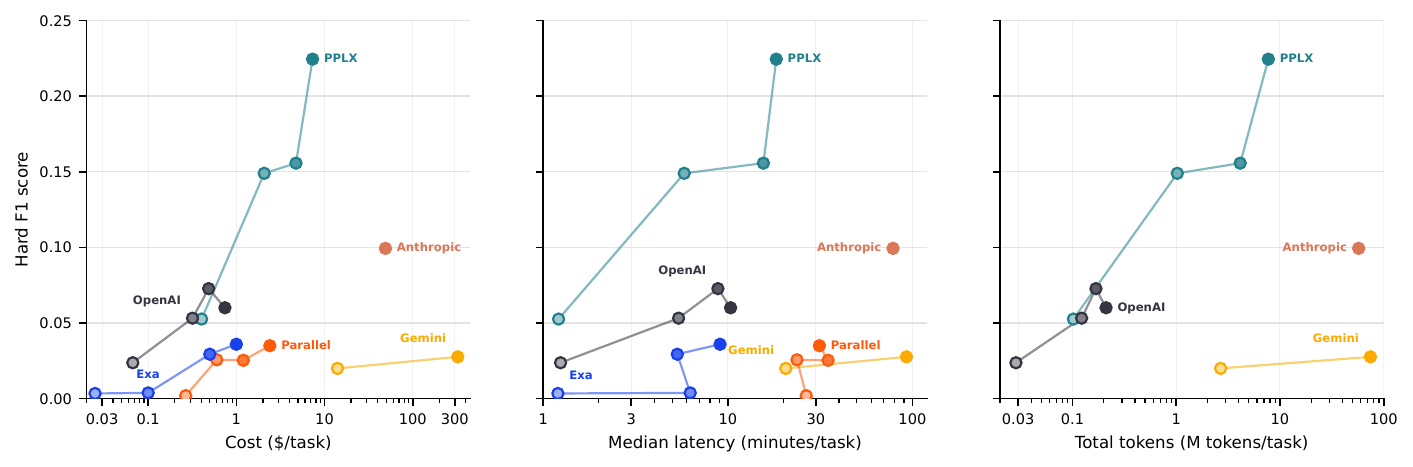}
        \caption{Hard F1.}
        \label{fig:sample-effort-hard-f1}
    \end{subfigure}
    \caption{F1 against cost, median solve latency, and total token use for the matched 45-task effort sweep in Tables~\ref{tab:supplemental-sweep-scores} and~\ref{tab:supplemental-sweep-operational}. Every point in (a) and (b) uses the same 45 tasks; for settings also run on the full benchmark, only the matched 45 tasks are included. System names directly label the colored paths, and darker color indicates higher effort. Exa and Parallel are absent from the token columns because their results expose no token counters.}
    \label{fig:sample-effort-f1}
\end{figure}

Taken together, Tables~\ref{tab:supplemental-sweep-scores} and~\ref{tab:supplemental-sweep-operational} and Figure~\ref{fig:sample-effort-f1} extend the comparison across all available effort settings. Perplexity, Gemini, and Exa improve monotonically in soft and hard F1 across their available settings; Perplexity \path{xhigh} reaches $0.447$ soft F1 and $0.224$ hard F1. OpenAI peaks at \path{high}, while Parallel's soft F1 rises through \path{ultra8x} but its hard F1 dips slightly between \path{ultra2x} and \path{ultra4x}. Resource use spans more than four orders of magnitude in cost, from Exa \path{low} at \$0.03 per task to Gemini \path{max} at \$324.83.

\clearpage
\begin{table}[H]
\centering
\small
\begin{tabular}{|cc|c|ccc|ccc|}
\hline
\multirow{2}{*}{\textbf{System}} & \multirow{2}{*}{\textbf{Setting}} & \multirow{2}{*}{\textbf{Completed}} & \multicolumn{3}{c|}{\textbf{Soft}} & \multicolumn{3}{c|}{\textbf{Hard}} \\
& & & \textbf{Precision} & \textbf{Recall} & \textbf{F1} & \textbf{Precision} & \textbf{Recall} & \textbf{F1} \\
\hline
Perplexity & \path{low}    & \textbf{45} & 0.214 & 0.109 & 0.121 & 0.098 & 0.051 & 0.053 \\
Perplexity & \path{medium} & \textbf{45} & 0.350 & 0.289 & 0.295 & \underline{0.166} & 0.156 & 0.149 \\
Perplexity & \path{high}   & \textbf{45} & \underline{0.414} & \underline{0.391} & \underline{0.397} & 0.162 & \underline{0.160} & \underline{0.156} \\
Perplexity & \path{xhigh}  & \underline{44} & \textbf{0.459} & \textbf{0.449} & \textbf{0.447} & \textbf{0.242} & \textbf{0.249} & \textbf{0.224} \\
\hline
Anthropic  & \path{high} & \textbf{45} & 0.365 & 0.238 & 0.262 & 0.155 & 0.095 & 0.099 \\
\hline
OpenAI     & \path{low}    & \textbf{45} & 0.106 & 0.032 & 0.038 & 0.027 & 0.024 & 0.024 \\
OpenAI     & \path{medium} & \textbf{45} & 0.158 & 0.087 & 0.091 & 0.067 & 0.055 & 0.053 \\
OpenAI     & \path{high}   & \textbf{45} & 0.177 & 0.150 & 0.153 & 0.076 & 0.081 & 0.073 \\
OpenAI     & \path{xhigh}  & \textbf{45} & 0.148 & 0.128 & 0.127 & 0.063 & 0.069 & 0.060 \\
\hline
Gemini     & \path{speed}  & \textbf{45} & 0.127 & 0.036 & 0.048 & 0.068 & 0.015 & 0.020 \\
Gemini     & \path{max}    & \textbf{45} & 0.106 & 0.069 & 0.074 & 0.049 & 0.025 & 0.028 \\
\hline
Parallel   & \path{ultra}   & 40 & 0.013 & 0.004 & 0.005 & 0.005 & 0.002 & 0.002 \\
Parallel   & \path{ultra2x} & \textbf{45} & 0.156 & 0.050 & 0.063 & 0.069 & 0.021 & 0.026 \\
Parallel   & \path{ultra4x} & \textbf{45} & 0.179 & 0.053 & 0.067 & 0.077 & 0.020 & 0.025 \\
Parallel   & \path{ultra8x} & \textbf{45} & 0.170 & 0.068 & 0.080 & 0.075 & 0.032 & 0.035 \\
\hline
Exa        & \path{low}    & \textbf{45} & 0.136 & 0.004 & 0.006 & 0.128 & 0.002 & 0.003 \\
Exa        & \path{medium} & \textbf{45} & 0.054 & 0.009 & 0.013 & 0.025 & 0.003 & 0.004 \\
Exa        & \path{high}   & \textbf{45} & 0.151 & 0.058 & 0.073 & 0.058 & 0.026 & 0.029 \\
Exa        & \path{xhigh}  & \textbf{45} & 0.142 & 0.103 & 0.111 & 0.039 & 0.035 & 0.036 \\
\hline
\end{tabular}
\caption{Matched 45-task score sweep across all available system settings. Every score is a simple zero-filled mean over the same 45 scheduled tasks. Bold marks the best value in each numeric column and underline marks the second-best distinct value, using unrounded values.}
\label{tab:supplemental-sweep-scores}
\end{table}

\begin{table}[H]
\centering
\small
\begin{tabular}{|cc|c|c|cc|cc|}
\hline
\multirow{2}{*}{\textbf{System}} & \multirow{2}{*}{\textbf{Setting}} & \multirow{2}{*}{\textbf{Completed}} & \multicolumn{1}{c|}{\textbf{Cost}} & \multicolumn{2}{c|}{\textbf{Latency}} & \multicolumn{2}{c|}{\textbf{Token usage}} \\
& & & \textbf{\$/task} & \textbf{Med. min} & \textbf{P90 min} & \textbf{In M} & \textbf{Out k} \\
\hline
Perplexity & \path{low}    & \textbf{45} & 0.40 & \underline{1.2} & \underline{3.4} & \underline{0.10} & \textbf{3.3} \\
Perplexity & \path{medium} & \textbf{45} & 2.07 & 5.8 & 19.3 & 1.01 & 13.7 \\
Perplexity & \path{high}   & \textbf{45} & 4.75 & 15.6 & 36.9 & 4.11 & 28.2 \\
Perplexity & \path{xhigh}  & \underline{44} & 7.32 & 18.3 & 32.3 & 7.69 & 43.2 \\
\hline
Anthropic  & \path{high} & \textbf{45} & 49.09 & 78.5 & 132.7 & 57.04 & 316.6 \\
\hline
OpenAI     & \path{low}    & \textbf{45} & \underline{0.07} & 1.2 & \textbf{2.1} & \textbf{0.02} & \underline{4.3} \\
OpenAI     & \path{medium} & \textbf{45} & 0.32 & 5.4 & 8.0 & 0.10 & 19.5 \\
OpenAI     & \path{high}   & \textbf{45} & 0.49 & 8.8 & 13.8 & 0.14 & 32.1 \\
OpenAI     & \path{xhigh}  & \textbf{45} & 0.74 & 10.4 & 14.4 & 0.16 & 55.3 \\
\hline
Gemini     & \path{speed}  & \textbf{45} & 14.03 & 20.6 & 58.3 & 2.60 & 83.1 \\
Gemini     & \path{max}    & \textbf{45} & 324.83 & 92.9 & 322.7 & 73.62 & 896.1 \\
\hline
Parallel   & \path{ultra}   & 40 & 0.27 & 26.6 & 55.3 & -- & -- \\
Parallel   & \path{ultra2x} & \textbf{45} & 0.60 & 23.7 & 37.8 & -- & -- \\
Parallel   & \path{ultra4x} & \textbf{45} & 1.20 & 34.9 & 66.2 & -- & -- \\
Parallel   & \path{ultra8x} & \textbf{45} & 2.40 & 31.3 & 58.1 & -- & -- \\
\hline
Exa        & \path{low}    & \textbf{45} & \textbf{0.03} & \textbf{1.2} & 5.9 & -- & -- \\
Exa        & \path{medium} & \textbf{45} & 0.10 & 6.3 & 11.5 & -- & -- \\
Exa        & \path{high}   & \textbf{45} & 0.50 & 5.3 & 7.2 & -- & -- \\
Exa        & \path{xhigh}  & \textbf{45} & 1.00 & 9.1 & 17.1 & -- & -- \\
\hline
\end{tabular}
\caption{Operational statistics for the same matched 45-task sweep as Table~\ref{tab:supplemental-sweep-scores}. Cost and token totals are divided by 45 scheduled tasks; median and P90 are solve-stage latencies over tasks with valid timestamps. In M and Out k denote millions and thousands of tokens. Exa and Parallel expose no token counters. Bold marks the best value and underline marks the second-best distinct value, using unrounded values; higher is better for completion and lower is better for resource use.}
\label{tab:supplemental-sweep-operational}
\end{table}

\subsection{Failure Analysis} \label{sec:failure-analysis}

\subsubsection{WANDR as a Retrieval Program}

Solving a WANDR task amounts to executing a retrieval program over a required entity space: discover enough members, enrich every required branch, qualify the resulting pages, and render source-backed records. We analyze four observable components of that program. \emph{Breadth--depth--extract execution} carries candidates from discovery through enrichment to evidence extraction while tracking quotas and backfilling sparse branches. \emph{Identity disambiguation} canonicalizes keys, collapses duplicates, and attaches evidence to the correct entity. \emph{Semantic qualification} checks type, eligibility, source role, and page-level support. \emph{Evidence rendering} selects faithful, sufficient excerpts and emits normalized records. The layered verifier makes these components inspectable at structural, task, and record resolution.

Search as Code (SaC) is well matched to this repeated horizontal structure. It lets a model compose retrieval, ranking, filtering, fan-out, and rendering primitives through generated Python, while the sandbox executes batching, retries, joins, aggregation, and deduplication without a separate model turn for every web operation. Filesystem state can preserve candidate tables and quota deficits across turns, supporting systematic backfilling instead of repeated serial search-and-read cycles \citep{perplexity2026sac}. Perplexity's observed profile is consistent with that advantage: it has the smallest soft precision--recall gap, the highest post-discovery retention, and the strongest conditional evidence completion without the highest cost, latency, or token use (Tables~\ref{tab:results-scores}, \ref{tab:results-operational}, and~\ref{tab:evidence-rendering}; Figure~\ref{fig:retrieval-decomposition}).

\subsubsection{Breadth--Depth--Extract Execution}

\paragraph{Precision--recall gaps expose missing breadth} Under \path{verdict_full}, soft recall is lower than soft precision for every system. Precision averages the quality of submitted members, whereas recall also collapses duplicate identities and zero-pads any shortfall against the required member count. Perplexity retains $0.357$ recall from $0.389$ precision, the smallest gap; Anthropic falls from $0.354$ to $0.222$, the largest. Because measured identity collapse changes endpoint-average full soft recall by at most $0.127$ percentage points, under-delivery accounts for most of these gaps (Tables~\ref{tab:results-scores} and~\ref{tab:identity-collapse}).

\paragraph{Soft--hard gaps expose incomplete members} Soft metrics reward partial subtrees; hard metrics credit only members whose required descendants and checks are all complete. Perplexity falls from $0.363$ soft F1 to $0.133$ hard F1, Anthropic from $0.249$ to $0.072$, and the remaining systems from $0.055$--$0.121$ to $0.009$--$0.035$. The loss also appears in precision---$0.389$ to $0.150$ for Perplexity and $0.354$ to $0.137$ for Anthropic---so many submitted members are only partially complete. SaC leads every hard metric, but coordinated identity, qualification, and rendering failures still prevent full-member completion (Table~\ref{tab:results-scores}).

\begin{figure}[h]
    \centering
    \includegraphics[width=\linewidth]{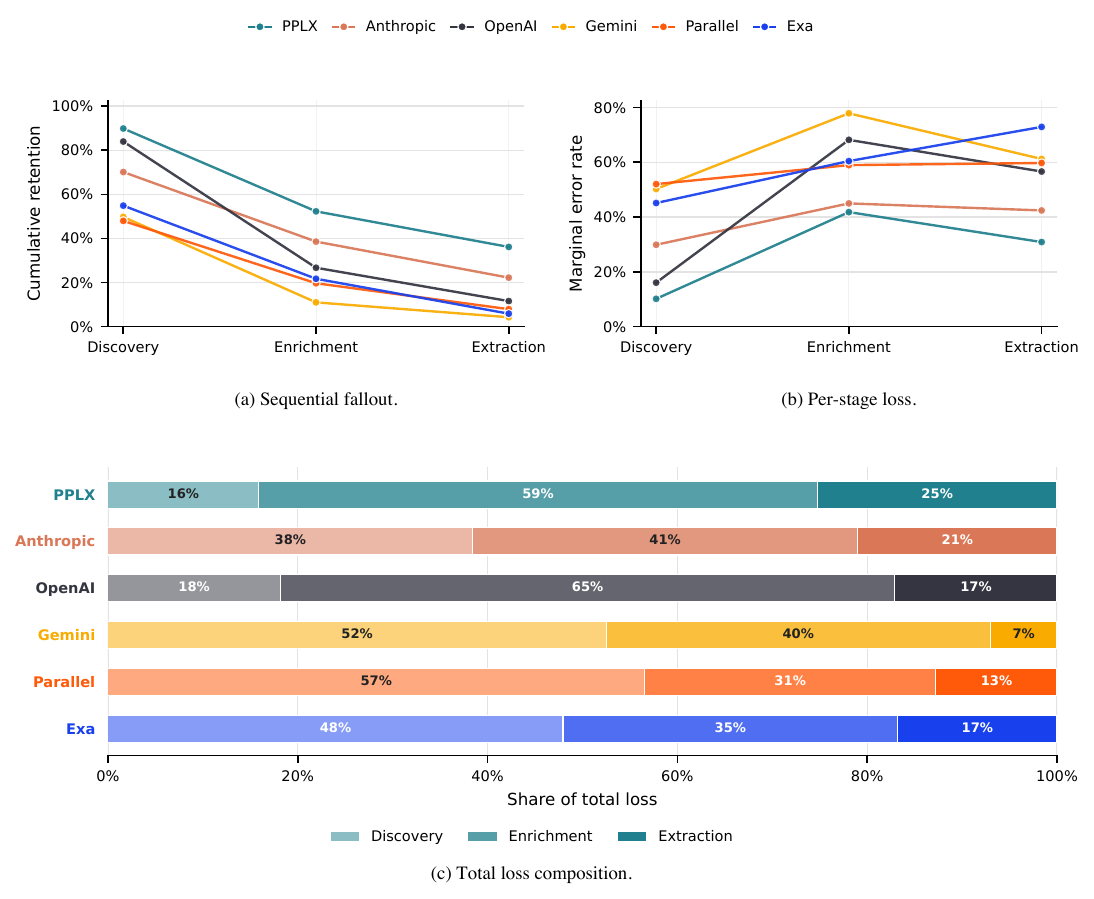}
    \caption{Decomposition of full soft recall into discovery, enrichment, and extraction after factoring out recognized identity losses. (a) Cumulative retention after each stage. (b) Error at each stage conditional on reaching it. (c) Each stage's share of the total loss; segments sum to $100\%$ within an endpoint. The small correction described in the text aligns the reconstructed product with reported full soft recall. Endpoint order is Perplexity, Anthropic, OpenAI, Gemini, Parallel, and Exa, based on 496, 500, 498, 493, 350, and 490 scored submissions, respectively.}
    \label{fig:retrieval-decomposition}
\end{figure}

\paragraph{Task structure separates discovery, enrichment, and extraction} We factor full soft recall into three successive retentions over the same submitted members. \emph{Discovery} is quota-retained delivered members divided by required members; \emph{enrichment} is their mean composed retrieval-only soft recall; and \emph{extraction} is their summed full soft recall divided by summed retrieval-only soft recall. The product reconstructs full soft recall after recognized identity effects are separated, making stage-local and cumulative losses directly comparable (Figure~\ref{fig:retrieval-decomposition}).\footnote{A small endpoint-level correction ($0.01\%$--$0.31\%$) reconciles the reconstructed product with reported full soft recall; restoring the identity effect in Table~\ref{tab:identity-collapse} then reproduces the reported score exactly.}

\paragraph{Late extraction cliffs do not dominate total loss} Extraction removes $30.9\%$--$72.9\%$ of the score that reaches it, yet contributes only $7.0\%$--$25.3\%$ of total loss and is the smallest component for five systems. Discovery and enrichment have already removed $47.7\%$--$89.0\%$ of the initial opportunity. OpenAI illustrates the converse: it retains $83.9\%$ through discovery, then loses $68.2\%$ of the remainder during enrichment, making enrichment $64.7\%$ of its total loss (Figure~\ref{fig:retrieval-decomposition}).

\begin{figure}[h]
    \centering
    \includegraphics[width=\linewidth]{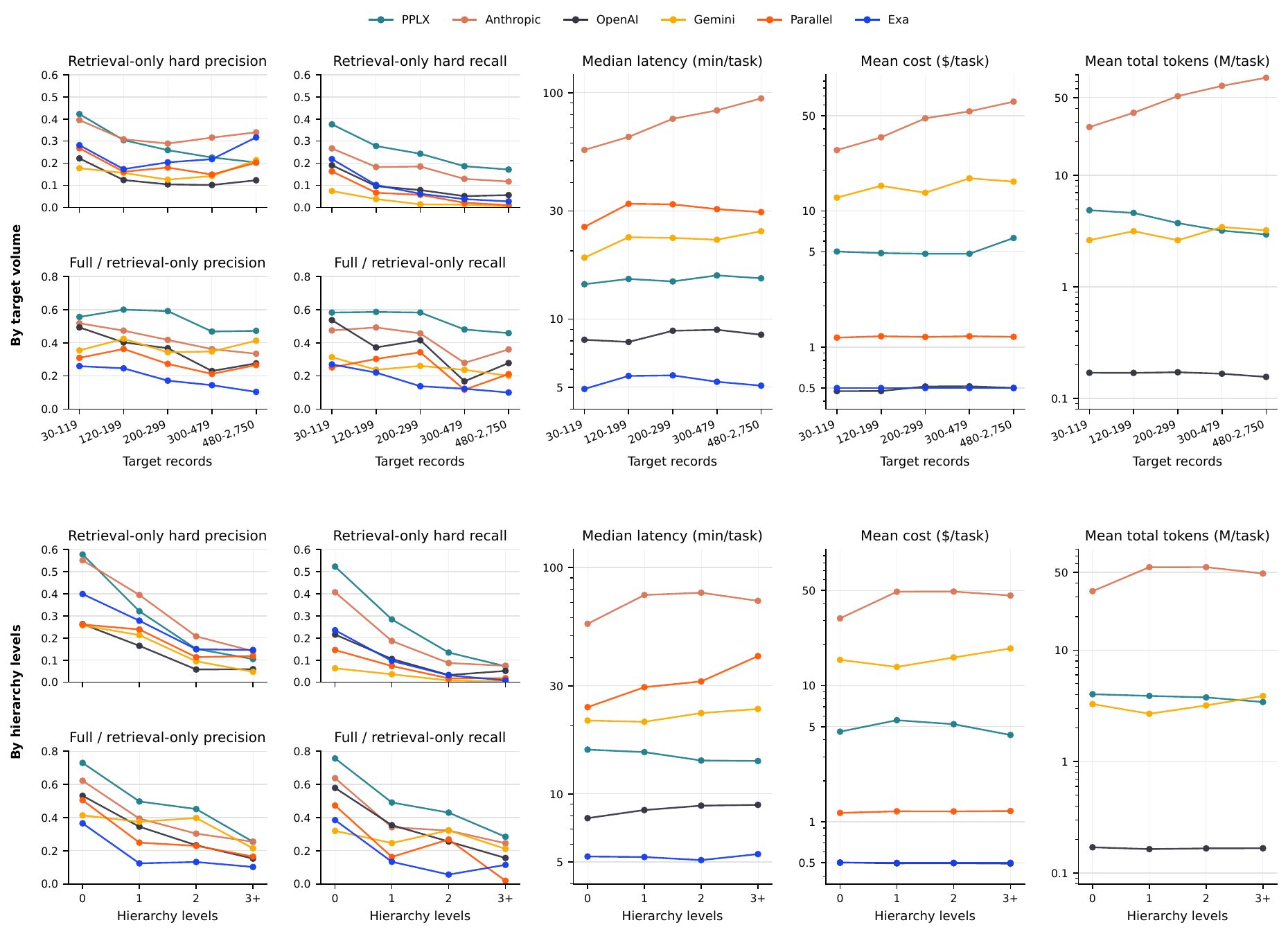}
    \caption{Task-level hard precision, hard recall, and resource scaling across target record volume (top block) and number of hierarchy levels (bottom block). Within each block, the upper-left panels show retrieval-only scores and the lower-left panels show full-to-retrieval conversion. Each conversion is the bin's mean full score divided by its mean retrieval-only score, so multiplying vertically reconstructs the mean full score exactly. The shared resource panels span both score rows because operational statistics are unchanged. Each line is one fixed main-run system configuration; points are equal-task aggregates within bins. Scores, mean solver cost, and mean provider-reported total tokens use the scheduled bin denominator, with missing values zero-filled to match the benchmark aggregate within systems that expose each measure; latency is the median over tasks with valid solve timestamps. Exa and Parallel are absent from the token panels because they expose no token counters. The target-volume bins contain 87, 103, 86, 117, and 107 tasks, and the hierarchy-level groups contain 64, 225, 158, and 53 tasks; $3+$ pools tasks with three to six distinct non-member, non-URL keys across the root and all subtasks. Cost, latency, and token use have log y-axes.}
    \label{fig:scaling-analysis}
\end{figure}

\paragraph{Target volume correlates primarily with pre-extraction erosion} From the smallest to largest required-volume bin, retrieval-only hard recall falls from $0.376$ to $0.172$ for Perplexity, $0.267$ to $0.117$ for Anthropic, and $0.191$ to $0.056$ for OpenAI. Full-to-retrieval conversion weakens less consistently: the first-to-last decline averages $10.4$ percentage points for precision and $13.7$ for recall across systems, while Gemini's precision conversion rises. Resource use does not explain a common scaling response: Anthropic's latency, cost, and token use rise sharply, but Perplexity's token use falls from $4.87$M to $2.95$M and OpenAI's remains near $0.17$M (Figure~\ref{fig:scaling-analysis}, top block).

\paragraph{More hierarchy levels compound both losses} From zero to three or more hierarchy levels, Perplexity's retrieval-only hard precision falls from $0.578$ to $0.105$ and hard recall from $0.523$ to $0.072$; Anthropic falls from $0.551$ to $0.141$ and from $0.407$ to $0.074$; and OpenAI falls from $0.263$ to $0.059$ and from $0.215$ to $0.051$. Unlike target volume, additional levels reduce both precision and recall conversion for every system, by $33.7$ and $35.3$ percentage points on average from the shallowest to deepest bin. Resource use again has no shared monotonic response, so these observational curves localize the loss but do not isolate hierarchy as its cause (Figure~\ref{fig:scaling-analysis}, bottom block).

\FloatBarrier

\subsubsection{Identity Disambiguation}

\paragraph{Identity collapse is sparse globally but material when it changes score} The scorer merges submitted values that refer to the same entity, preventing aliases from inflating coverage. Comparing full soft recall with and without the merge penalty captures both lost duplicate count and the take-worst penalty when variants of one identity carry evidence of different quality (Table~\ref{tab:identity-collapse}).

\begin{table}[H]
\centering
\small
\begin{tabular}{@{}lrrrrr@{}}
\toprule
\multirow{2}{*}{\textbf{System}} & \multirow{2}{*}{\textbf{Tasks}} & \multicolumn{2}{c}{\textbf{Duplicate collapse}} & \multicolumn{2}{c}{\textbf{Full soft-recall loss (pp)}} \\
& & \textbf{Affected (\%)} & \textbf{Values (\%)} & \textbf{All tasks} & \textbf{When nonzero} \\
\midrule
Perplexity & 500 & 4.8 & 0.426 & 0.127 & 4.52 \\
Anthropic  & 500 & 2.4 & 0.018 & 0.032 & 1.98 \\
OpenAI     & 499 & 7.0 & 0.353 & 0.067 & 2.43 \\
Gemini     & 493 & 5.5 & 0.086 & 0.017 & 0.77 \\
Parallel   & 350 & 8.0 & 0.184 & 0.036 & 0.78 \\
Exa        & 490 & 6.3 & 0.191 & 0.046 & 2.05 \\
\bottomrule
\end{tabular}
\caption{Identity collapse in the six full runs. Affected is the share of tasks with at least one merged entity identity; Values is the share of submitted entity-key values merged into another identity. All tasks is the mean full soft-recall loss; When nonzero is the mean conditional on identity collapse lowering recall. The performance columns use the scored submissions available for 496, 500, 498, 493, 350, and 490 tasks in displayed order. One OpenAI submission moves in the opposite direction and remains in the all-task net effect.}
\label{tab:identity-collapse}
\end{table}

Across systems, only $2.4\%$--$8.0\%$ of tasks contain a duplicate entity identity, and only $0.018\%$--$0.426\%$ of submitted entity values collapse. The mean full soft-recall effect is therefore small, $0.017$--$0.127$ percentage points, but rises to $0.77$--$4.52$ points when collapse changes a task's score. Identity is not a major aggregate explanation of endpoint differences, but it can materially damage an affected task (Table~\ref{tab:identity-collapse}; Appendix~\ref{appendix:worked-scoring}).

\paragraph{SaC concentrates rather than eliminates identity errors} Perplexity has duplicate identities on only $4.8\%$ of tasks, below Gemini, Exa, OpenAI, and Parallel, but it has both the largest collapsed-value share ($0.426\%$) and the largest conditional soft-recall loss ($4.52$ points). Parallel has the highest affected-task share ($8.0\%$), yet one of the smallest conditional effects ($0.78$ points). SaC's identity failures are therefore less frequent than in four systems but more concentrated when they occur (Table~\ref{tab:identity-collapse}).

\subsubsection{Semantic Qualification}

\begin{figure}[h]
    \centering
    \includegraphics[width=\linewidth]{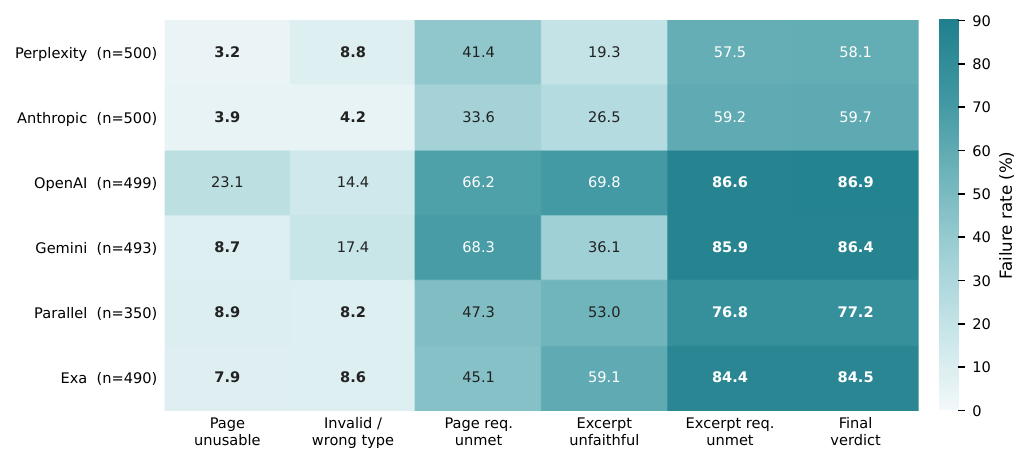}
    \caption{Measured row-level failure probes from available verifier details (lower is better). Page req. unmet means that the fetched page fails at least one substantive requirement; Excerpt req. unmet means that the submitted excerpts fail at least one. Final verdict denotes failure under \texttt{verdict\_full}. Each cell reports the failure rate in percent, rounded to one decimal place; row labels give the number of available task details. Coverage is 500, 500, 499, 493, 350, and 490 tasks for Perplexity, Anthropic, OpenAI, Gemini, Parallel, and Exa, respectively; Parallel's field analysis is therefore less complete than its aggregate results.}
    \label{fig:failure-fields}
\end{figure}

\paragraph{Usable, valid pages are common; complete page support is not} For five systems, only $3.2\%$--$8.9\%$ of submitted pages are unusable; OpenAI is the $23.1\%$ outlier (Figure~\ref{fig:failure-fields}). Invalid-or-wrong-type failures are also comparatively limited at $4.2\%$--$17.4\%$. The larger problem begins after a page is fetched and judged plausible: $33.6\%$--$68.3\%$ of records cite a page that does not satisfy every substantive task requirement. Anthropic performs best on this check at $33.6\%$ failure, with Perplexity second at $41.4\%$. Finding an accessible, broadly in-scope page is therefore not the main semantic bottleneck; finding a page that establishes the complete requested claim is.

\FloatBarrier

\subsubsection{Evidence Rendering}

\paragraph{Faithful excerpts usually support something, but not everything} We condition on records whose fetched page satisfies every substantive requirement, then measure whether the submitted excerpts are faithful, whether a faithful excerpt supports at least one requirement, and whether it supports all requirements (Table~\ref{tab:evidence-rendering}).

\begin{table}[H]
\centering
\small
\begin{tabular}{@{}lrrr@{}}
\toprule
\multirow{2}{*}{\textbf{System}} & \multirow{2}{*}{\textbf{Faithful (\%)}} & \multicolumn{2}{c}{\textbf{Given faithful (\%)}} \\
& & \textbf{Any supported} & \textbf{All supported} \\
\midrule
Perplexity & 85.6 & 99.4 & 80.4 \\
Anthropic  & 83.1 & 99.2 & 70.0 \\
OpenAI     & 57.1 & 99.3 & 75.7 \\
Gemini     & 79.1 & 96.9 & 55.6 \\
Parallel   & 62.9 & 99.1 & 71.0 \\
Exa        & 49.2 & 98.4 & 59.7 \\
\bottomrule
\end{tabular}
\caption{Evidence rendering after conditioning on records whose fetched page satisfies all substantive requirements. Values are equal-task means. Faithful uses all such records; the two support columns then condition on faithful excerpts. Any supported is computed from the task-specific support checks; All supported uses the judge's aggregate all-requirements field. The initial condition is observed on 494, 499, 489, 476, 341, and 473 tasks in displayed order; the conditional columns have 488, 497, 464, 470, 336, and 457 tasks with at least one faithful record.}
\label{tab:evidence-rendering}
\end{table}

\paragraph{Partial support is the dominant rendering failure} Once an excerpt is faithful, it supports something on $96.9\%$--$99.4\%$ of records, but supports every requirement on only $55.6\%$--$80.4\%$. The characteristic failure is therefore a faithful, partly useful selection that omits facts needed to verify the complete record. OpenAI has low faithfulness ($57.1\%$) but relatively strong completion once faithful ($75.7\%$), whereas Gemini has high faithfulness ($79.1\%$) but the weakest complete support ($55.6\%$; Table~\ref{tab:evidence-rendering}).

\begin{figure}[h]
    \centering
    \includegraphics[width=\linewidth]{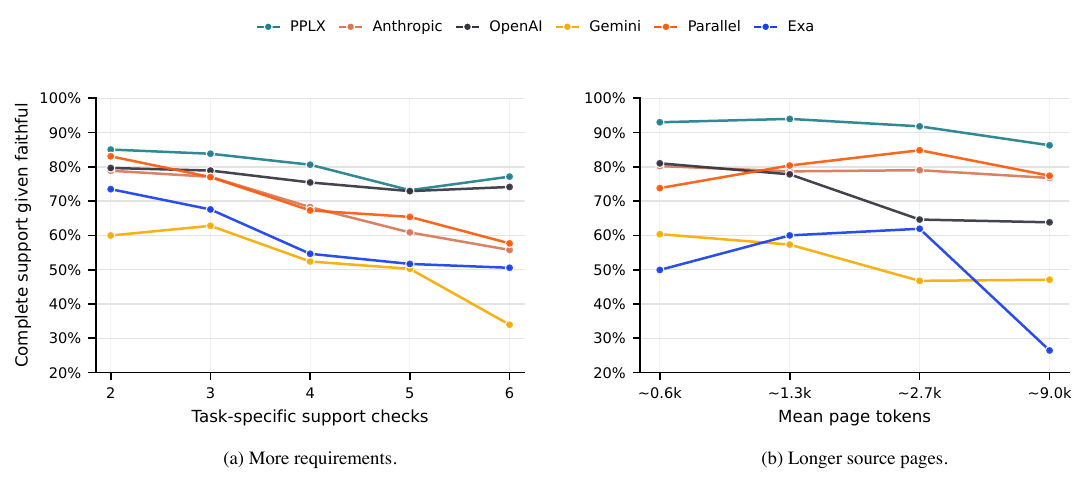}
    \caption{Complete excerpt support after conditioning on a page that satisfies all requirements and on faithful excerpts. (a) Equal-task rates by the number of task-specific support checks in the full runs. (b) Rates by within-task fetched-page-length quartile in a deterministic 12-task-per-system cache sample. Tick labels give the mean page length in each quartile, estimated at four characters per token. The page-length analysis covers 7,193 of 7,339 page-satisfied records with joined content.}
    \label{fig:evidence-rendering}
\end{figure}
\FloatBarrier

\paragraph{More requirements widen the evidence-selection gap} Complete support falls between tasks with two task-specific support checks and those with five for all six systems, by $6.8$--$21.8$ percentage points. This association is descriptive because requirement count also varies with task content and structure, but it consistently exposes a broader selection burden after the page and excerpt pass the preceding checks (Figure~\ref{fig:evidence-rendering}(a)).

\paragraph{Long pages show the same selective-completeness problem} In the cache sample, complete support falls from the shortest to longest within-task page-length quartile in five of six systems. Manual inspection matches the pattern: failures omit a date from a ranking page, air-burden language from a permit, or an individual review from a long review surface even when the selected text is faithful and partly relevant. This analysis is exploratory because page length co-varies with source type and record complexity (Figure~\ref{fig:evidence-rendering}(b)).

\subsubsection{Cross-System Synthesis}

\paragraph{Failures compound across layers} Recall falls below precision, hard scores fall far below soft scores, and performance deteriorates as target volume and hierarchy levels increase. The stage decomposition localizes the largest share of soft-recall loss to enrichment for Perplexity, Anthropic, and OpenAI, but to discovery for Gemini, Parallel, and Exa. Downstream, substantive page support fails much more often than page usability or basic validity, and faithful excerpts frequently omit part of the required proof (Table~\ref{tab:results-scores}; Figures~\ref{fig:retrieval-decomposition}, \ref{fig:scaling-analysis}, and~\ref{fig:failure-fields}; Table~\ref{tab:evidence-rendering}).

\paragraph{SaC's advantage is broad but not uniform} Perplexity combines the strongest breadth retention with the highest conditional evidence completion, but Anthropic has the lowest page-requirement failure and Perplexity has the largest conditional identity loss. OpenAI is the page-usability and excerpt-faithfulness outlier; Gemini is usually faithful but has the weakest complete support once faithful; and Parallel has the highest identity-affected task share, although its diagnostic coverage is limited to 350 tasks. SaC's distinctive signature is therefore stronger breadth retention and evidence construction, not uniformly better performance at every layer (Figure~\ref{fig:failure-fields}; Tables~\ref{tab:identity-collapse} and~\ref{tab:evidence-rendering}).

\section{Discussion}\label{sec:discussion}
\subsection{Summary}
WANDR evaluates a form of research common in professional work but poorly represented by single-answer and report-writing benchmarks: constructing a large structured collection while independently verifying every member. Its 500 tasks combine a large-scale, often open-ended breadth requirement with a per-record depth requirement, and its qualification key hierarchy makes discovery, enrichment, and evidence failures separately observable. Reference-free grading accommodates changing facts without maintaining an exhaustive gold answer table; each submitted record instead carries the page and excerpt needed for verification.

The evaluation shows that this problem remains far from solved. Across six production systems, Perplexity Search as Code leads at $0.363$ soft F1 and $0.133$ hard F1, but even its hard recall is only $0.134$. Perplexity has the smallest soft precision--recall gap and the strongest evidence-rendering funnel, the profile expected from programmable fan-out, batching, filtering, joining, deduplication, and target-aware backfilling. Its depth performance is not uniformly best, however. Anthropic has the lowest page-requirement failure rate, and even for Perplexity only $80.4\%$ of faithful excerpts completely support the claim after conditioning on a page that does. The common bottleneck is therefore not search alone but complete evidence construction at scale.

The four-layer framework measures failures at aggregate or task-local resolution. Breadth--depth--extract execution produces both precision--recall and soft--hard gaps; within soft recall, discovery, enrichment, and extraction show where continuous coverage is lost. Complete performance also drops sharply with larger targets and more hierarchy levels. Identity disambiguation has little aggregate effect but can cost several full soft-recall points on affected submissions. In semantic qualification, page usability and basic validity are usually not the main failures; substantive page support is. Within evidence rendering, the dominant signature is partial proof: faithful excerpts usually support at least one requirement but often not all of them. More generally, the results support treating wide research as a systems problem coupling retrieval, deterministic data processing, identity management, semantic qualification, state management, and evidence rendering.

Even though WANDR is primarily an evaluation benchmark, the same structure makes it a plausible RL substrate. The task pipeline can generate held-out training packages, per-level required counts support curriculum learning, record-level verdicts and hierarchy-level scores provide denser outcome rewards than a single exact match, and the streaming grader can amortize overlapping work across batched rollouts. Appendix~\ref{appendix:rl-training} discusses this design and its caveats, including the need to train on generated or separately held-out tasks rather than the released benchmark split.

\subsection{Limitations}
\paragraph{Scope and representativeness} The 500 released tasks are designed to represent recurring patterns in de-identified production requests, but they are not a random sample of all research work. Admission deliberately favors high-volume, feasible, discriminative tasks with independently verifiable web evidence. This makes WANDR useful as a stress test, but the resulting score distribution should not be interpreted as the difficulty of research requests in deployment.

\paragraph{System comparison and causal attribution} The main evaluation reports one production run per system rather than repeated trials to save time and cost. An earlier, separately configured evaluation reported the same qualitative ordering \citep{perplexity2026sac}, but it also used one run per system and therefore does not estimate run-to-run variance. Systems differ simultaneously in model, search infrastructure, agent harness, tool interface, and provider-side implementation. The cross-system ranking is therefore observational and does not isolate any individual component.

\paragraph{Reference-free grading} Evidence verification avoids a manual and static gold table but introduces judge and retrieval uncertainty. The grader uses LLM judge calls and re-fetches each cited page: the page may drift after solving, and extraction can omit dynamic, tabular, or JavaScript-rendered content. Canonicalization and deduplication introduce additional boundary decisions when identities are ambiguous. We keep solving and grading close in time, and retry broken pages with a browser, but these controls reduce rather than eliminate grading error.

\paragraph{Agentic fetch backend alignment} The grader first re-fetches cited pages through an agentic fetch backend that is also available to the Perplexity solver; unsuccessful fetches are routed to a browser. Because this backend is shared, the grading path may be better aligned with pages that the Perplexity solver can access than with pages reached through other systems' retrieval stacks. In internal grading ablations that disable the agentic fetch backend and use only the browser, scores decrease for every system while the ranking remains mostly stable. The grading fetch path therefore affects absolute scores and may confer some alignment advantage, even though the comparative ordering appears less sensitive. Future evaluations should report the grading fetch configuration and compare multiple retrieval paths.

\subsection{Future Research Directions}
\paragraph{Controlled scaling studies} Figure~\ref{fig:scaling-analysis} identifies descriptive scale sensitivity in the current fixed-configuration runs, but target volume, intermediate-level count, content, and difficulty co-vary. A useful next experiment is a matched comparison that varies one factor at a time: workload scale, model, search interface, sandbox, skills, or retrieval infrastructure.

\paragraph{Mechanism-resolved failure analysis} The failure analysis is observational: score trees and verifier fields identify where submitted outputs lose credit, but they do not establish which model, retrieval, orchestration, or interface mechanism caused the loss. Controlled ablations should hold the task sample and the rest of the system fixed while varying one component at a time, such as code execution, parallel fan-out, target-aware backfilling, context persistence, identity handling, or the fetcher. Paired runs with repeated trials would distinguish causal effects from task composition and run-to-run variation.

\paragraph{RL training experiments} Appendix~\ref{appendix:rl-training} explains how WANDR can serve as a training environment; the next step is to test the claim. Experiments can compare terminal F1 alone against denser, record- and branch-level rewards, measure whether quota curricula improve performance, and quantify the grading savings from shared queues and caches as task-by-rollout batch size grows. Training should use generated sibling tasks or newly generated tasks from appropriate held-out seeds, preserve a held-out benchmark split, and test for reward hacking---especially over-submission, evidence templating, and exploitation of judge or cache artifacts.

\subsection{Conclusion}
WANDR makes large-scale, evidence-backed research measurable without an exhaustive gold collection. Across 500 tasks, recall falls below precision, hard scores remain far below soft scores, and complete performance degrades sharply as target volume and the number of hierarchy levels increase. Page usability and basic validity usually succeed, but substantive page support and especially complete excerpt support remain major bottlenecks. It provides a basis for improving comprehensive search systems and testing whether those gains hold at scale.

\clearpage
\begingroup
\bibliographystyle{plainnat}
\bibliography{references}

@inproceedings{mialon2023gaia,
  title={{GAIA}: A Benchmark for General {AI} Assistants},
  author={Mialon, Gr{\'e}goire and Fourrier, Cl{\'e}mentine and Swift, Craig and Wolf, Thomas and LeCun, Yann and Scialom, Thomas},
  booktitle={The Twelfth International Conference on Learning Representations},
  year={2024}
}

@article{phan2025humanity,
  title={Humanity's Last Exam},
  author={Phan, Long and Gatti, Alice and Han, Ziwen and Li, Nathaniel and Hu, Josephina and Zhang, Hugh and Zhang, Chen Bo Calvin and Shaaban, Mohamed and Ling, John and Shi, Sean and others},
  journal={arXiv preprint arXiv:2501.14249},
  year={2025}
}

@inproceedings{krishna2024frames,
  title={Fact, Fetch, and Reason: A Unified Evaluation of Retrieval-Augmented Generation},
  author={Krishna, Satyapriya and Krishna, Kalpesh and Mohananey, Anhad and Schwarcz, Steven and Stambler, Adam and Upadhyay, Shyam and Faruqui, Manaal},
  booktitle={Proceedings of the 2025 Conference of the Nations of the Americas Chapter of the Association for Computational Linguistics: Human Language Technologies (Volume 1: Long Papers)},
  pages={4745--4759},
  year={2025},
  doi={10.18653/v1/2025.naacl-long.243},
  url={https://aclanthology.org/2025.naacl-long.243/}
}

@inproceedings{yoran2024assistantbench,
  title={{AssistantBench}: Can Web Agents Solve Realistic and Time-Consuming Tasks?},
  author={Yoran, Ori and Amouyal, Samuel Joseph and Malaviya, Chaitanya and Bogin, Ben and Press, Ofir and Berant, Jonathan},
  booktitle={Proceedings of the 2024 Conference on Empirical Methods in Natural Language Processing},
  pages={8938--8968},
  year={2024},
  doi={10.18653/v1/2024.emnlp-main.505},
  url={https://aclanthology.org/2024.emnlp-main.505/}
}

@article{wei2025browsecomp,
  title={{BrowseComp}: A Simple Yet Challenging Benchmark for Browsing Agents},
  author={Wei, Jason and Sun, Zhiqing and Papay, Spencer and McKinney, Scott and Han, Jeffrey and Fulford, Isa and Chung, Hyung Won and Passos, Alex Tachard and Fedus, William and Glaese, Amelia},
  journal={arXiv preprint arXiv:2504.12516},
  year={2025}
}

@inproceedings{gou2025mind2web,
  title={{Mind2Web 2}: Evaluating Agentic Search with Agent-as-a-Judge},
  author={Gou, Boyu and Huang, Zanming and Ning, Yuting and Gu, Yu and Lin, Michael and others},
  booktitle={Advances in Neural Information Processing Systems, Datasets and Benchmarks Track},
  year={2025}
}

@article{chen2025medbrowsecomp,
  title={MedBrowseComp: Benchmarking Medical Deep Research and Computer Use},
  author={Chen, Shan and Moreira, Pedro and Xiao, Yuxin and Schmidgall, Sam and Warner, Jeremy and Aerts, Hugo and Hartvigsen, Thomas and Gallifant, Jack and Bitterman, Danielle S},
  journal={arXiv preprint arXiv:2505.14963},
  year={2025}
}

@article{zhu2025findeepresearch,
  title={{FinDeepResearch}: Evaluating Deep Research Agents in Rigorous Financial Analysis},
  author={Zhu, Fengbin and Ng, Xiang Yao and Liu, Ziyang and Liu, Chang and Zeng, Xianwei and Wang, Chao and Tan, Tianhui and Yao, Xuan and Shao, Pengyang and Xu, Min and others},
  journal={arXiv preprint arXiv:2510.13936},
  year={2025}
}

@inproceedings{li2025legalagentbench,
  title={{LegalAgentBench}: Evaluating {LLM} Agents in Legal Domain},
  author={Li, Haitao and Chen, Junjie and Yang, Jingli and Ai, Qingyao and Jia, Wei and Liu, Youfeng and Lin, Kai and Wu, Yueyue and Yuan, Guozhi and Hu, Yiran and others},
  booktitle={Proceedings of the 63rd Annual Meeting of the Association for Computational Linguistics (Volume 1: Long Papers)},
  pages={2322--2344},
  year={2025},
  doi={10.18653/v1/2025.acl-long.116},
  url={https://aclanthology.org/2025.acl-long.116/}
}

@article{zhou2025academicbrowse,
  title={{ScholarSearch}: Benchmarking Scholar Searching Ability of {LLM}s},
  author={Zhou, Junting and Li, Wang and Liao, Yiyan and Zhang, Nengyuan and Miao, Tingjia and Qi, Zhihui and Wu, Yuhan and Yang, Tong},
  journal={arXiv preprint arXiv:2506.13784},
  year={2025}
}

@article{bigeard2025finance,
  title={Finance Agent Benchmark: Benchmarking {LLM}s on Real-World Financial Research Tasks},
  author={Bigeard, Antoine and Nashold, Langston and Krishnan, Rayan and Wu, Shirley},
  journal={arXiv preprint arXiv:2508.00828},
  year={2025}
}

@article{gupta2026deepsearchqa,
  title={DeepSearchQA: Bridging the Comprehensiveness Gap for Deep Research Agents},
  author={Gupta, Nikita and Chatterjee, Riju and Haas, Lukas and Tao, Connie and Wang, Andrew and Liu, Chang and Oiwa, Hidekazu and Gribovskaya, Elena and Ackermann, Jan and Blitzer, John and others},
  journal={arXiv preprint arXiv:2601.20975},
  year={2026}
}

@article{lan2025deepwidesearch,
  title={{DeepWideSearch}: Benchmarking Depth and Width in Agentic Information Seeking},
  author={Lan, Tian and Zhu, Bin and Jia, Qianghuai and Ren, Junyang and Li, Haijun and Wang, Longyue and Xu, Zhao and Luo, Weihua and Zhang, Kaifu},
  journal={arXiv preprint arXiv:2510.20168},
  year={2025}
}

@article{xiong2026autoresearchbench,
  title={{AutoResearchBench}: Benchmarking {AI} Agents on Complex Scientific Literature Discovery},
  author={Xiong, Lei and Luo, Kun and Xia, Ziyi and Zhang, Wenbo and Yao, Jin-Ge and Liu, Zheng and Shao, Jingying and Chen, Jianlyu and Qian, Hongjin and others},
  journal={arXiv preprint arXiv:2604.25256},
  year={2026}
}

@article{huang2026wideseek,
  title={{WideSeek}: Advancing Wide Research via Multi-Agent Scaling},
  author={Huang, Ziyang and Ren, Haolin and Yuan, Xiaowei and Wang, Jiawei and Jiang, Zhongtao and Xu, Kun and He, Shizhu and Zhao, Jun and Liu, Kang},
  journal={arXiv preprint arXiv:2602.02636},
  year={2026}
}

@article{zhu2026gisa,
  title={{GISA}: A Benchmark for General Information-Seeking Assistant},
  author={Zhu, Yutao and Zhang, Xingshuo and Zhang, Maosen and Jin, Jiajie and Zhang, Liancheng and Song, Xiaoshuai and Zhao, Kangzhi and Zeng, Wencong and Tang, Ruiming and Li, Han and Wen, Ji-Rong and Dou, Zhicheng},
  journal={arXiv preprint arXiv:2602.08543},
  year={2026}
}

@article{liu2025veriweb,
  title={{VeriWeb}: Verifiable Long-Chain Web Benchmark for Agentic Information-Seeking},
  author={Liu, Shunyu and Liu, Minghao and Zhou, Huichi and Cui, Zhenyu and Zhou, Yang and Zhou, Yuhao and Gao, Jialiang and Zhou, Heng and Yang, Yunhao and others},
  journal={arXiv preprint arXiv:2508.04026},
  year={2025}
}

@article{lan2026tableassearch,
  title={{Table-as-Search}: Formulate Long-Horizon Agentic Information Seeking as Table Completion},
  author={Lan, Tian and Henry, Felix and Zhu, Bin and Jia, Qianghuai and Ren, Junyang and Pu, Qihang and Li, Haijun and Wang, Longyue and Xu, Zhao and Luo, Weihua},
  journal={arXiv preprint arXiv:2602.06724},
  year={2026}
}

@article{zhuge2024agent,
  title={{Agent-as-a-Judge}: Evaluate Agents with Agents},
  author={Zhuge, Mingchen and Zhao, Changsheng and Ashley, Dylan and Wang, Wenyi and Khizbullin, Dmitrii and Xiong, Yunyang and Liu, Zechun and Chang, Ernie and Krishnamoorthi, Raghuraman and Tian, Yuandong and others},
  journal={arXiv preprint arXiv:2410.10934},
  year={2024}
}

@misc{anthropic2026managedagents,
  author       = {{Anthropic}},
  title        = {Claude Managed Agents Overview},
  howpublished = {\url{https://platform.claude.com/docs/en/managed-agents/overview}},
  year         = {2026},
  note         = {Accessed: 2026-07-13}
}

@misc{perplexity2026sac,
  title        = {Rethinking Search as Code Generation},
  author       = {{Perplexity}},
  year         = {2026},
  url          = {https://research.perplexity.ai/articles/rethinking-search-as-code-generation},
  note         = {Perplexity Research. Accessed: 2026-06-19}
}

@article{zhong2026draco,
  title        = {{DRACO}: A Cross-Domain Benchmark for Deep Research Accuracy, Completeness, and Objectivity},
  author       = {Zhong, Joey and Zhang, Hao and Southern, Clare and Yang, Jeremy and Wang, Thomas and Jung, Kate and Zhang, Shu and Yarats, Denis and Ho, Johnny and Ma, Jerry},
  year         = {2026},
  journal      = {arXiv preprint arXiv:2602.11685},
  url          = {https://arxiv.org/abs/2602.11685}
}

@article{wong2025widesearch,
  title        = {{WideSearch}: Benchmarking Agentic Broad Info-Seeking},
  author       = {Wong, Ryan and Wang, Jiawei and Zhao, Junjie and others},
  journal      = {arXiv preprint arXiv:2508.07999},
  year         = {2025},
  note         = {ByteDance Seed}
}
\endgroup

\clearpage
\appendix
\renewcommand{\thesubsection}{\Alph{subsection}}

\section*{Appendices}

\subsection{Per-Record Grading Criteria} \label{appendix:criteria}

Criteria define what one submitted record must prove. The judge evaluates each leaf independently from the submitted item, answer, URL, and excerpts, together with the fetched page.

\subsubsection{Layered Checks}

Each record is evaluated through universal checks, task-specific validity checks, and substantive requirements. Universal checks apply to every task. Validity checks decide whether the record is eligible and well-formed enough to judge. Requirements state the facts that the cited evidence must establish.

Every substantive requirement is evaluated twice. A field ending in \path{_satisfied} asks whether the full page supports the requirement, regardless of which excerpts the solver selected. Its paired \path{_supported} field asks whether the submitted excerpts alone are sufficient for a careful reader to verify the same requirement. The schema aggregates these pairs into \path{requirements_all_satisfied} and \path{requirements_all_supported}.

In the running CEO/CFO example, the appointment page must identify the company and appointee, establish the role, and place the first public announcement in the target window. A page can satisfy those requirements even when the solver submits only an excerpt naming the appointee. In that case, the page-level fields pass, but the excerpt-level date or company fields fail because the submitted evidence is incomplete.

Three universal checks guard this requirement pair:
\begin{itemize}
    \item \path{page_content_usable} requires a reachable, substantive, on-topic page; dead links, stubs, and off-topic pages fail here.
    \item \path{answer_intent_clear} requires the record to state the specific claim being made. A page dump with no localized answer, or excerpts that contradict one another, fails here.
    \item \path{excerpts_faithful} requires every excerpt to appear verbatim or near-verbatim on the page with its meaning preserved. Paraphrase, fabricated text, misleading truncation, and sentence stitching fail this check.
\end{itemize}

Faithfulness and completeness are separate. An excerpt may quote the page exactly but omit the sentence that establishes the date, role, or source authority; it then passes \path{excerpts_faithful} but fails the relevant \path{_supported} field. Conversely, text stitched from separate passages may mention every needed fact but fail faithfulness because the page never states the combined claim. This separation is especially important when a solver relies on search snippets, which can omit qualifiers or join nearby text out of context.

Task-specific validity checks cover eligibility conditions that do not fit a page/excerpt requirement pair. Some are structural, such as a required excerpt length or allowed domain; others are broader sanity checks, such as whether a submitted person or company belongs to the requested class. The schema summarizes them in \path{overall_valid}. If a record is so invalid that substantive judgment would be meaningless, the judge short-circuits the remaining fields to false; otherwise, it still reports the requirement-level signals for diagnosis.

\begin{figure}[h]
    \centering
    \includegraphics[width=\linewidth]{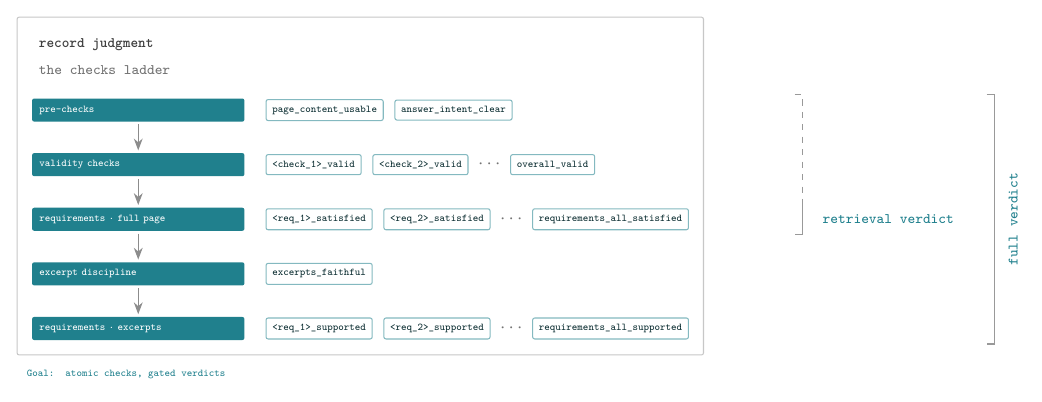}
    \caption{The record-level checks. \texttt{verdict\_retrieval} reads the full-page requirement signal, shown by the solid bracket segment; its dashed continuation shows the preceding checks that the judge may use to gate satisfiability when relevant, not a mechanistic conjunction. \texttt{verdict\_full} also requires a usable page, a clear answer, faithful excerpts, complete excerpt support, and task-specific validity. A task-specific validity check can short-circuit an ineligible record.}
    \label{fig:checks-ladder}
\end{figure}

Figure~\ref{fig:checks-ladder} summarizes the resulting sequence. As defined in Section~\ref{sec:grading}, \path{verdict_retrieval} asks whether the fetched page satisfies every substantive task requirement, independently of the submitted excerpts. Universal and validity fields are not mechanically multiplied into this verdict, but the judge may use them to gate page-level satisfiability when relevant. The complete-record verdict, \path{verdict_full}, additionally requires validity, a usable page, a clear answer, faithful excerpts, and complete excerpt support. These are the two record-level verdicts aggregated by the task metrics.

\subsubsection{Common Evidence Patterns}

The exact requirements vary by task, but several evidence patterns recur across the benchmark:

\begin{itemize}
    \item \emph{Plausible entity} --- the submitted entity must be a sensible instance of the requested class. A quote author should be a real person rather than a fictional character; an LLM producer should build models rather than merely resell access to them.
    \item \emph{Verified eligibility} --- some restrictions are key-level validity checks, while others require cited evidence. In the running example, the appointment branch checks that the company is US-based, while the listing subtask requires listing-authority evidence of public status.
    \item \emph{Appropriate source} --- the citation must come from a source suited to the claim, such as an official filing, first-party product page, recognized listing authority, or practitioner forum. The allowed source type can be established by domain, page authorship, or in-page evidence.
    \item \emph{Long-tail source} --- the qualifying evidence may sit outside the highest-ranked results. These tasks reward systematic search beyond prominent pages and aggregators.
    \item \emph{Heavy source} --- the evidence may live in a long report, registry, filing, or PDF. Verifying the claim requires fetching the artifact and navigating to the relevant passage rather than relying on a snippet.
    \item \emph{Dedicated source} --- the page must be substantially about the submitted entity, not a directory row, search-result page, or passing mention. For example, a task may require a dedicated product listing rather than a buyer's guide that names the product once.
    \item \emph{Broad extraction} --- the claim itself may require many pieces of evidence. Showing that a governing body has at least $N$ members, for example, requires excerpts that enumerate enough members and therefore yields empirically long evidence blocks even without an explicit word-count floor. A task can also impose such a floor directly.
    \item \emph{Substantive extraction} --- the page and excerpts must contain analysis or detail, not merely a rating, label, or short listing. A narrative-review requirement filters out bare aggregate scores.
    \item \emph{Scattered extraction} --- the required evidence appears in different parts of a page, such as a title, date, and substantive comment. The solver must collect all of it without joining unrelated passages.
    \item \emph{Structured extraction} --- exact values sit in tables, filings, or specification sheets. The main risk is reading the wrong cell, row, or column; GPU--game FPS results are a typical example.
\end{itemize}

\clearpage
\subsection{Identity Resolution} \label{appendix:identity}

Identity rules determine when two submitted values count as the same member. Without them, cosmetic variants could satisfy a volume floor more than once: ``Aleksandr Pushkin'' and ``Alexander Pushkin'' should not count as two authors, nor should ``NVIDIA'' and ``Nvidia Corp.''\ count as two companies.

Leaf verdicts remain independent, but scoring must compare values across the submission to count distinct members. WANDR limits this cross-record step to one sameness policy per key axis. Once identity is resolved on each axis, required volumes, coverage, and score rollups are well defined.

\subsubsection{Canonicalization and Deduplication}

WANDR uses two complementary mechanisms:

\begin{itemize}
    \item \emph{Canonicalization} maps a submitted value to a standard form. It is well suited to closed or predictable sets: ``Mar. 2025'' and ``March 2025'' can both map to ``2025-03,'' dispatch labels can map to a fixed vocabulary, and URLs can be normalized mechanically. A closed-set canonicalizer can also reject values outside the allowed set.
    \item \emph{Deduplication} decides whether two values refer to the same entity. Exact matching is sufficient after deterministic normalization for fields such as URLs or fixed labels. Open-ended entities such as people and companies usually require semantic comparison.
\end{itemize}

Both mechanisms are configured per key axis. Each key must contain enough information to identify its values without relying on an implicit parent. Composite keys supply that context when one field is ambiguous. In the running example, \texttt{company\_appointee} includes both the company and appointee fields, distinguishing appointments of the same person---or same-named people---across different companies. A single-field key is sufficient when its values are already unambiguous within the task.

When a subtask reuses a parent key, both branches share one identity axis. In the running example, company names from the appointment and listing branches are resolved together before the two branch scores are composed.

\subsubsection{Common Identity Patterns}

Identity policy is part of the task definition, not merely cleanup. Changing what counts as the same entity changes the set the task asks the solver to cover.

\begin{itemize}
    \item \emph{Granularity} --- the policy sets the level being counted. Folding franchise locations and subsidiaries into one identity creates a brand-level task; keeping them separate creates an outlet- or company-level task. A closed set may deliberately mix levels when the task treats them as peers, such as a roster containing states, the District of Columbia, and selected county-level regimes.
    \item \emph{One referent across forms} --- renames, abbreviations, and different stages of the same event can map to one underlying member. A product can remain the same offering after a rebrand, and coverage of a transaction's announcement and closing can refer to the same transaction even though the pages use different language.
    \item \emph{Exact-set coverage} --- a finite canonical set changes ``find at least $n$'' into ``cover every listed value.'' Aliases map to the same allowed member, and values outside the solver-facing roster do not satisfy the axis.
\end{itemize}

\clearpage
\subsection{Task Package Format} \label{appendix:package-format}

Each released WANDR task is a self-contained package. Complete package files, prompts, fixtures, and verifier code live in the public repository.

\begin{table}[H]
\centering
\small
\renewcommand{\arraystretch}{1.12}
\begin{tabular}{@{}>{\raggedright\arraybackslash}p{0.20\linewidth}>{\raggedright\arraybackslash}p{0.39\linewidth}>{\raggedright\arraybackslash}p{0.33\linewidth}@{}}
\toprule
\textbf{Component} & \textbf{What it contains} & \textbf{Why it matters} \\
\midrule
Solver-facing task text & The natural-language request shown to a system under test. & Defines the agent's obligation without exposing the judge checklist as a separate answer key. \\
Key hierarchy & The tree of identifying keys ending in \texttt{url}, with required child counts at each level. & Determines the member unit, breadth, depth, zero-padding, and where a failure is localized. \\
Volume floors & Per-level \texttt{required} counts and, where relevant, closed sets or open ``all that qualify'' settings. & Enforces coverage without enumerating a complete gold answer table. \\
Judge schema & Universal fields plus task-specific validity gates and requirement pairs. & Specifies what the verifier checks on the cited page and in the submitted excerpts. \\
Identity policy & Canonicalization for closed-form keys and deduplication policy for open-ended identities. & Prevents duplicate padding and resolves surface variants of the same member. \\
Calibration fixtures & Small records with intended verdicts, including positive and negative cases. & Unit-tests the task's judge before the task can enter the benchmark set. \\
Metadata and labels & Structural labels, distribution labels, and release bookkeeping. & Supports dataset statistics, curation, and reproducible benchmark filtering. \\
Verifier harness & Shared fetch, judge, identity-resolution, and scoring code pinned with the release. & Lets outside users regrade submissions under the same task-level mechanics, subject to live-web and judge-model dependencies. \\
\bottomrule
\end{tabular}
\caption{WANDR task package format.}
\label{tab:package-format}
\end{table}

\clearpage
\subsection{Task Admission Gates and Review} \label{appendix:admission-details}

The authoring prompts and linter instructions are implementation artifacts in the release repository. A task must be runnable, convention-clean, sufficiently difficult, discriminative, empirically feasible, and judgeable before it can enter the released set.

\begin{table}[H]
\centering
\small
\renewcommand{\arraystretch}{1.12}
\begin{tabular}{@{}>{\raggedright\arraybackslash}p{0.20\linewidth}>{\raggedright\arraybackslash}p{0.55\linewidth}>{\raggedright\arraybackslash}p{0.17\linewidth}@{}}
\toprule
\textbf{Gate} & \textbf{Check} & \textbf{Failure action} \\
\midrule
Runs & The package loads and executes under the public harness with its fixtures. & Return to task or fixture authoring \\
Clean & The solver-facing description, judge specification, and schema state the same requirements in the same order; mechanical lints pass. & Patch mechanical issues or return to authoring \\
Difficulty & Primitive internal baselines do not already achieve high completion. & Harden the task or reject an easy seed \\
Discriminative & Weak internal baselines score low while stronger internal rollouts score meaningfully higher. & Tighten task or reject shortcut-prone seed \\
Feasibility audit & A merged solution over 10--12 authoring rollouts recovers a near-full requested volume floor. & Adjust volume or rewrite scope \\
Judge audit & A reviewer blindly regrades sampled records and attributes disagreements to solver behavior, task design, or grader machinery. & Fix judge or block task \\
Human rubric & Where automatic checks do not settle the decision, reviewers assess the key hierarchy, task text, judge, and identity handling. & Return to owning stage \\
\bottomrule
\end{tabular}
\caption{Admission and review gates for a task package. Curation follows admission and selects from the admitted pool to match release-level distributions.}
\label{tab:admission-gates}
\end{table}

\begin{table}[H]
\centering
\small
\renewcommand{\arraystretch}{1.12}
\begin{tabular}{@{}>{\raggedright\arraybackslash}p{0.24\linewidth}>{\raggedright\arraybackslash}p{0.64\linewidth}@{}}
\toprule
\textbf{Review axis} & \textbf{Substantive question} \\
\midrule
Key hierarchy & Are keys sensible, are volume floors achievable and nontrivial, and is URL corroboration neither missing nor excessive? \\
Task authorship & Does the solver-facing prose unambiguously entail the graded obligation without hidden requirements or shortcutable ambiguity? \\
Judge authorship & Does the judge schema encode the right validity gates and page/excerpt requirement checks? \\
Identity handling & Are canonicalization and deduplication configured for the kind of identity being counted? \\
\bottomrule
\end{tabular}
\caption{Human-review rubric.}
\label{tab:human-rubric}
\end{table}
Reviewers also inspect the rollouts and trajectories accumulated during authoring. This exposes degenerate solution strategies and shortcuts that may not be apparent from the package or aggregate scores alone, and lets reviewers return the task for hardening before admission.

\clearpage
\subsection{Empirical Task Statistics} \label{appendix:empirical-task-statistics}

The structural statistics in Section~\ref{sec:stats} describe what a task requires, but not how that requirement behaves in practice. Historical rollouts accreted as tasks were authored, validated, and hardened: authors used them to test feasibility, critics inspected them for shortcuts and ambiguous obligations, and reviewers used them to decide whether a task needed another revision. We aggregate that archive to characterize WANDR empirically at three resolutions. The task-ontology view measures where structured completion becomes difficult; the individual-citation view measures source reuse and diversity; and the trajectory view measures exploration and effort.

\begin{figure}[h]
    \centering
    \includegraphics[width=\linewidth]{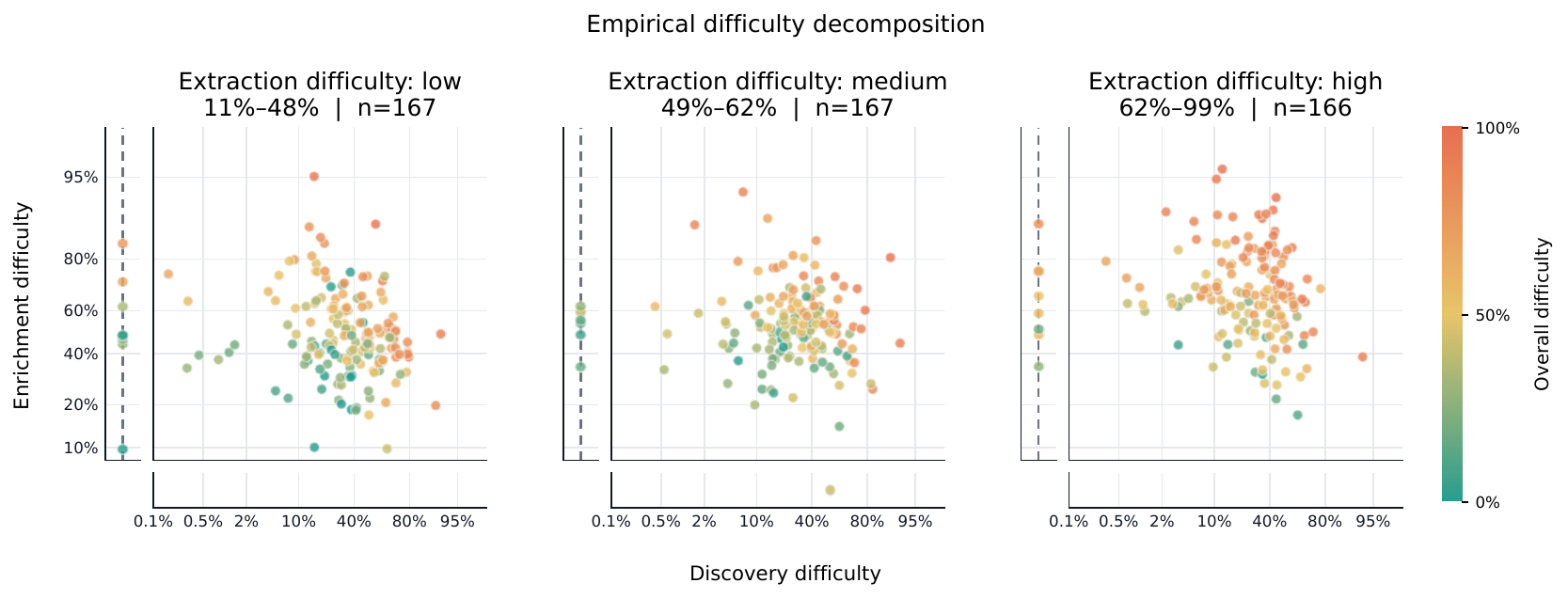}
    \caption{Task-ontology view of empirical difficulty over 500 tasks. Positive discovery and enrichment difficulty are shown on log-odds axes; exact-zero discovery values appear in the detached vertical strips and the single exact-zero enrichment value appears in the detached horizontal strip. Panels are equal-count extraction-difficulty tertiles. Color denotes overall difficulty.}
    \label{fig:empirical-difficulty}
\end{figure}

\paragraph{Task-ontology difficulty} Figure~\ref{fig:empirical-difficulty} decomposes structured attainment into discovery, enrichment, and extraction. For each rollout, discovery difficulty is one minus the delivered-member fraction; enrichment difficulty is one minus retrieval-only soft recall averaged over delivered members; and extraction difficulty is one minus the conversion from retrieval-only to full soft recall over the same members. The plotted task values are plain means over available rollouts, so unsuccessful executions are not downweighted. Overall difficulty is one minus the best soft F1 average observed during authoring.

The extraction tertiles separate visibly: tasks with high extraction difficulty also tend to have higher enrichment difficulty. Discovery and enrichment have a modest inverse association ($r=-0.185$ in log-odds space among positive pairs): when discovery returns fewer members, the delivered subset can be easier to enrich. Detached marginal strips retain the exact-zero cases: 24 tasks for discovery, one for enrichment, and none for both.

\paragraph{Citation volume and source ecology} A submitted collection can contain many records while repeatedly relying on the same pages. For each rollout, URL nonreuse is the number of unique cited URLs divided by the number of judged records. A value of one means that every record uses a distinct URL; a value of $0.1$ corresponds to roughly one unique URL per ten records. We average this ratio across available rollouts.

We define effective citation volume as required records multiplied by the square root of URL nonreuse. Counting every required record fully would ignore citation reuse, while counting only unique URLs would ignore the work required to attach and validate the same page against multiple records. The square root offers a middle ground that we found well behaved in practice. Domain dispersion is unique domains divided by unique URLs, averaged across the same rollouts.

\begin{figure}[h]
    \centering
    \includegraphics[width=.75\linewidth]{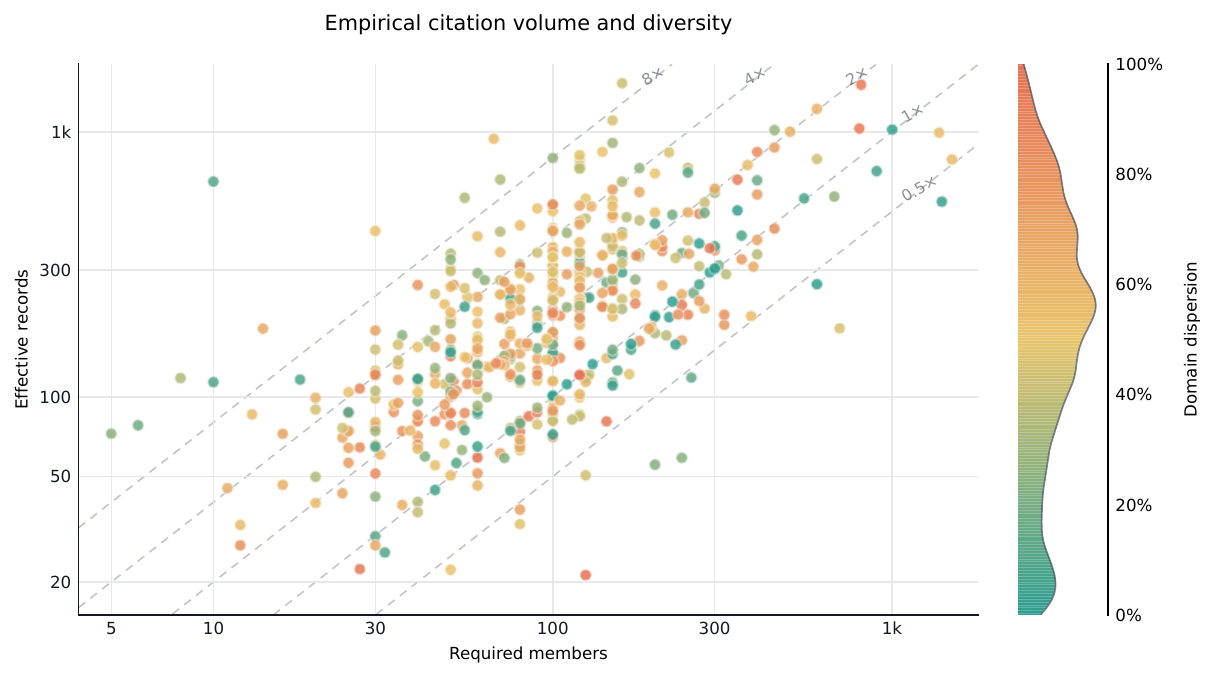}
    \caption{Individual-citation view of effective volume and source ecology over 500 tasks (log--log). Effective records equal required records times the square root of average URL nonreuse. Dashed guides mark constant effective records per required member. Point color denotes average domain dispersion; the sidebar shows its smoothed task distribution.}
    \label{fig:effective-citation-volume}
\end{figure}

Figure~\ref{fig:effective-citation-volume} places this effective volume against required members. The ratio guides therefore read as effective records per member. The median is $1.97$ effective records per member, with the central $90\%$ of tasks spanning $0.73$--$5.92$. Domain dispersion varies largely independently of scale: its median is $53.4\%$, while the 5th and 95th percentiles are $5.0\%$ and $86.9\%$. The distribution-shaped legend displays that concentration.

\begin{figure}[h]
    \centering
    \includegraphics[width=.75\linewidth]{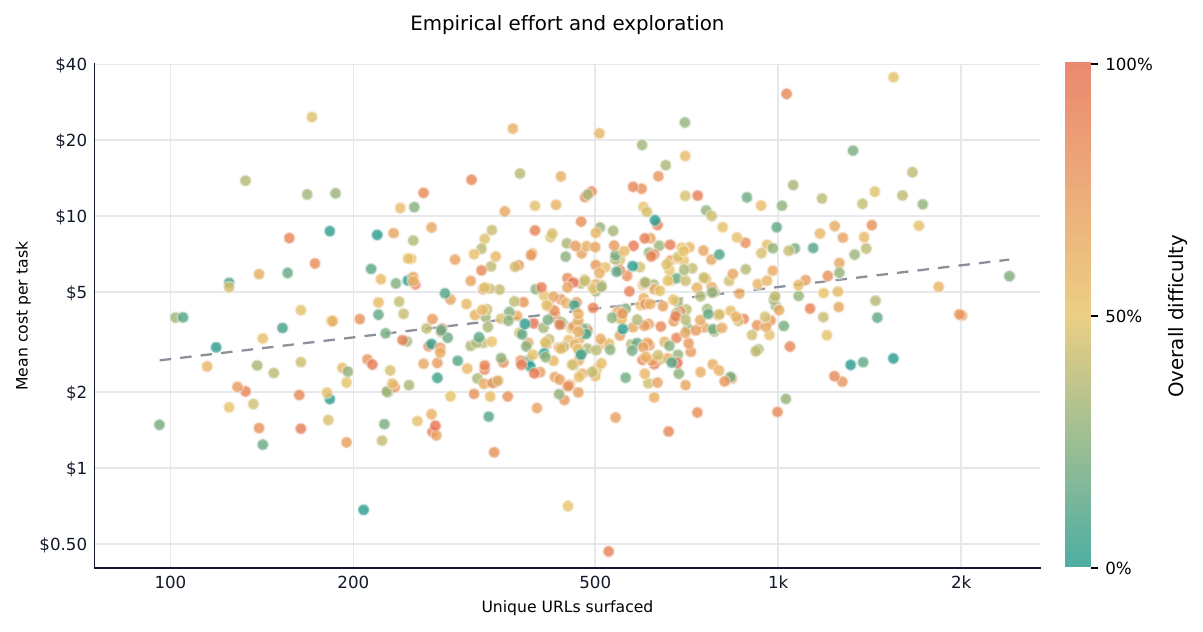}
    \caption{Trajectory view of exploration scope and effort over 490 non-stub authoring trajectories (log--log). Surfaced URLs include every unique URL touched through agent tool calls, not only submitted citations. Mean task cost is the geometric mean across available positive-cost rollouts. Point color denotes overall task difficulty; the dashed line is a least-squares fit in log space.}
    \label{fig:effort-exploration}
\end{figure}

\paragraph{Trajectory effort and execution} Citation counts describe the submitted collection, not the search process that produced it. We call a URL \emph{surfaced} when the agent interacted with it through any tool call, whether or not the URL appeared in the final submission. Figure~\ref{fig:effort-exploration} uses one consistently instrumented authoring trajectory per task and retains the same 490 non-stub trajectories used in the surfaced-URL distribution analysis; ten trajectories with fewer than 80 surfaced URLs are excluded. Task effort is the average cost across available rollouts. Because cost is multiplicative and strongly skewed, the mean is taken in log space and exponentiated.

The median retained trajectory surfaces 496 unique URLs, while the median task mean cost is \$4.00. Exploration scope and effort are positively but noisily associated ($r=0.280$ in log--log space).

\clearpage
\subsection{Exemplar Task Coverage} \label{appendix:paragons}

The authoring pipeline supplies a broader set of vetted guidance tasks to the author, critic, and linter. The five tasks profiled here are a deliberately varied subset that makes their design rationale explicit. They provide orientation, not templates: during design, the author can map a seed to a nearby exemplar archetype but may not copy its structure. Table~\ref{tab:exemplar-coverage} summarizes the distinct benchmark shapes covered by this profiled subset, followed by task cards describing the requested records and claims.

\begin{table}[H]
\centering
\scriptsize
\renewcommand{\arraystretch}{1.15}
\resizebox{\linewidth}{!}{%
\begin{tabular}{@{}>{\raggedright\arraybackslash}p{0.18\linewidth}>{\raggedright\arraybackslash}p{0.25\linewidth}>{\raggedright\arraybackslash}p{0.15\linewidth}>{\raggedright\arraybackslash}p{0.09\linewidth}>{\raggedright\arraybackslash}p{0.18\linewidth}>{\raggedright\arraybackslash}p{0.23\linewidth}@{}}
\toprule
\textbf{Exemplar} & \textbf{Key hierarchy sketch} & \textbf{Member unit} & \textbf{Records} & \textbf{Depth / identity pattern} & \textbf{Stress tested} \\
\midrule
\texttt{gpu\_benchmarks} & GPU $\rightarrow$ game $\rightarrow$ URL; subtasks for GPU price and game rating & GPU--game benchmark cell & 290 & Matrix fill with subtasks and corroborated enrichment; GPU/game deduplication and URL canonicalization & First-hand benchmark tables, gapless coverage, no second-hand roundup shortcut \\
\shortstack[l]{\texttt{bachir\_doha}\\\texttt{brand\_competitors}} & Brand $\rightarrow$ facet $\rightarrow$ URL & Brand--facet finding & 400 & Open-set discovery plus four-part dispatch; brand deduplication and facet canonicalization & Source-role discipline: review pages, menus, social pages, and positioning pages are not interchangeable \\
\shortstack[l]{\texttt{adult\_guardianship}\\\texttt{programs}} & Closed jurisdiction root plus legislation-signal subtask & Jurisdictional program or legal signal & 780 & Closed member panel with dispatch-like legal comparison areas; jurisdiction canonicalization & Legal authority, statutory fit, and multi-area enrichment over a fixed jurisdiction set \\
\shortstack[l]{\texttt{devtools}\\\texttt{partnerships}} & Company $\rightarrow$ partner $\rightarrow$ reference type $\rightarrow$ URL & Ordered partnership reference & 600 & Asymmetric pair identity with quote/backquote dispatch & Reciprocity, official-source control, and duplicate corporate-family traps \\
\texttt{llm\_presence} & Company $\rightarrow$ month $\rightarrow$ site $\rightarrow$ URL(2) & Company--month--site discussion cell & 720 & Temporal panel with site canonicalization and two-page corroboration per cell & Date anchoring, source diversity, and first-hand discussion filtering \\
\bottomrule
\end{tabular}
}
\caption{Coverage summary for the five exemplar tasks. The sketches intentionally abstract away the full package files, which are released in the repository.}
\label{tab:exemplar-coverage}
\end{table}

\begin{taskbox}[title={\texttt{gpu\_benchmarks}}]
\footnotesize
\textbf{Task:} For at least 20 GPUs and 10 games per GPU, find a source-backed FPS result (one URL per GPU--game pair) from a first-hand benchmark or review page. Two enrichment subtasks require each GPU's current retail price and each game's public review rating, with three URLs per item.\\[3pt]
\textbf{Record:} an FPS figure for one GPU--game pair, backed by a benchmark page.\\[3pt]
\textbf{Records required:} 290 (200 cells $+$ 60 price $+$ 30 review).\\[3pt]
\textbf{Claims per record:}\\
1.\ The benchmark is for the claimed GPU model --- \emph{proof: the benchmark page}\\
2.\ The benchmark is for the claimed game title, a real released game --- \emph{same page}\\
3.\ The page confirms a concrete FPS score --- \emph{same page}\\
4.\ Per GPU, its current retail price; per game, its review rating --- \emph{proof: three retailer or review pages per item (subtasks with corroboration)}
\end{taskbox}

\begin{taskbox}[title={\texttt{bachir\_doha\_brand\_competitors}}]
\footnotesize
\textbf{Task:} Identify 100+ dessert brands operating in Doha or elsewhere in Qatar (ice cream, gelato, frozen dessert, sweets, bakery dessert, or dessert caf\'e) and cover four analysis facets per brand---owned social identity, customer sentiment, delivery commerce, and market positioning---with one focused source per facet.\\[3pt]
\textbf{Record:} one facet finding for one brand, backed by a facet-appropriate page.\\[3pt]
\textbf{Records required:} 400 (100 brands $\times$ 4 facets).\\[3pt]
\textbf{Claims per record:}\\
1.\ The brand is a real, locally operating, in-scope Qatar brand --- \emph{validity}\\
2.\ The page ties the brand to the Qatar market --- \emph{proof: the facet page}\\
3.\ The page genuinely plays the claimed facet's source role (a review page cannot serve \emph{market positioning}) --- \emph{dispatch-routed check}\\
4.\ The page exposes a focused finding for that facet --- \emph{same page}
\end{taskbox}

\begin{taskbox}[title={\texttt{adult\_guardianship\_programs}}]
\footnotesize
\textbf{Task:} For all 30 listed jurisdictions, including the District of Columbia, identify the public or last-resort adult-guardianship arrangement (one URL each). A legislation subtask then finds five legal signals in each of five comparison areas per jurisdiction: appointment basis, decision-support alternatives, selection priority, oversight, and rights review.\\[3pt]
\textbf{Record (root):} a jurisdiction's institutional fallback arrangement, backed by a statute, agency, or program page.\\[3pt]
\textbf{Records required:} 780 (30 root $+$ $30 \times 5 \times 5$ subtask).\\[3pt]
\textbf{Claims per record:}\\
1.\ The named arrangement is the jurisdiction's institutional fallback for adults needing a guardian of last resort --- \emph{proof: the cited page}\\
2.\ The jurisdiction is one of the 30 items in the closed list --- \emph{closed-set validity}\\
3.\ (subtask) each legal signal's cited section fits its comparison area, on an authoritative legal surface --- \emph{proof: own page per signal, dispatch-specific}
\end{taskbox}

\begin{taskbox}[title={\texttt{devtools\_partnerships}}]
\footnotesize
\textbf{Task:} For at least 100 developer-tools or DevOps companies, find three advertised partners whose relationship is acknowledged in both directions. For every ordered (\texttt{company}, \texttt{other\_company}) pair, provide both a \emph{quote} (the company naming the partner) and a \emph{backquote} (the partner meaningfully acknowledging the company), with one URL for each direction.\\[3pt]
\textbf{Record:} one direction of one partnership---a page on the referencing party's own surface naming the counterpart.\\[3pt]
\textbf{Records required:} 600 ($100 \times 3 \times 2$).\\[3pt]
\textbf{Claims per record:}\\
1.\ The page is on a surface officially controlled by the referencing party --- \emph{proof: the page (incl.\ its host)}\\
2.\ It names the opposite party of the pair --- \emph{same page}\\
3.\ The reference is substantive for its type---a backquote must show meaningful reciprocity, not a logo wall or name-drop --- \emph{per-type check}\\
4.\ The counterpart is a genuine, distinct company (not an alias or the same corporate family) --- \emph{validity}
\end{taskbox}

\begin{taskbox}[title={\texttt{llm\_presence}}]
\footnotesize
\textbf{Task:} Identify 10+ LLM-producer companies and build a community-sentiment panel: for each company and each month in a fixed 12-month window, provide dedicated first-hand discussion or impression pages from three different sites, with two URLs per (company, month, site) cell.\\[3pt]
\textbf{Record:} one standalone community-discussion page about one company in one month on one site.\\[3pt]
\textbf{Records required:} 720 ($10 \times 12 \times 3 \times 2$).\\[3pt]
\textbf{Claims per record:}\\
1.\ The page is clearly dated to the target month (in-content dates, post timestamps, or thread markers) --- \emph{proof: the page}\\
2.\ It is on the claimed site --- \emph{same page (incl.\ host)}\\
3.\ It is a standalone, first-hand impression or discussion of the company---not a landing page, bulk thread view, or search-results display --- \emph{same page}\\
4.\ Two independent such pages per cell --- \emph{corroboration ($k{=}2$)}
\end{taskbox}

\clearpage
\subsection{Stylized Metric Rollup Algorithm} \label{appendix:metric-rollup}

Metric computation recursively follows the task structure (Section~\ref{sec:anatomy}). \emph{Criteria} supply the leaf score selected from the two verdicts defined in Section~\ref{sec:grading}. \emph{Topology} specifies which levels exist, what volume each requires, and where subtasks compose. \emph{Identity} determines which submitted values represent the same entity within each level. The rollup is parameterized by three choices: aggregation (\emph{precision} or \emph{recall}), continuity (\emph{soft} or \emph{hard}), and verdict. A subtask score is multiplied into the matching parent entity's score using the same choices and recursive procedure; a missing subtask value contributes zero.

\begin{pycode}
def score(task, level, identifiers, params):
    records = task.records.narrow(identifiers)
    if not records:
        return 0

    if level + 1 == len(task.keys):
        return records[0].judgment[params.verdict]  # leaf judgment
    next_key = task.keys[level + 1].name

    values = {r[next_key]: r["cluster_id"] for r in records}
    c_scores = [
        score(task, level + 1, identifiers | {next_key: v}, params)
        for v in values
    ]

    if params.aggregation == "recall":
        volume = task.keys[level + 1].volume
        clusters = duplicates(values, of=next_key)
        c_scores = pad(top(take_worst(c_scores, clusters), volume), 0, volume)
    s = mean(c_scores)

    if level == -1:  # thresholding and composition don't apply at overall-task avg
        return s

    if params.continuity == "hard":
        s = floor(s)

    curr_key = task.keys[level].name
    sub_scores = [
        score(sub, 0, {curr_key: identifiers[curr_key]}, params)
        for sub in task.subtasks(level)
    ]

    return s * product(sub_scores)

metric = score(root_task, -1, {}, params)
\end{pycode}
\captionof*{figure}{Stylized metric rollup. The released task packages include the production implementation.}

\clearpage
\subsection{Worked Rollup Walkthrough} \label{appendix:worked-scoring}

\begin{figure}[h]
    \centering
    \includegraphics[width=0.72\linewidth]{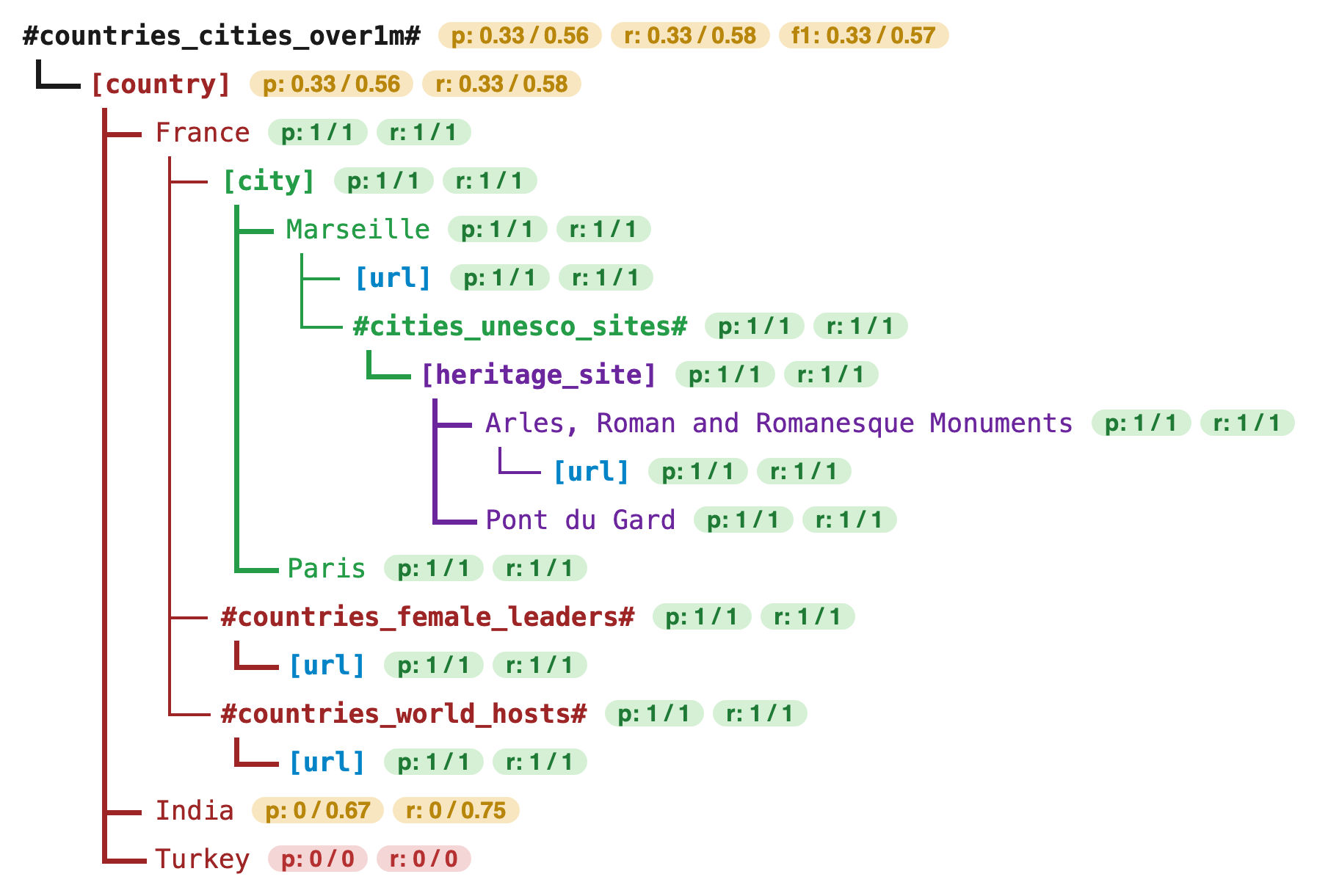}
    \caption{Overall score for the toy composite task. France is complete, India receives partial credit, and Turkey fails a required subtask, yielding soft F1 $0.57$ and hard F1 $0.33$.}
    \label{fig:toy-overview}
\end{figure}

\begin{figure}[h]
    \centering
    \includegraphics[width=\linewidth]{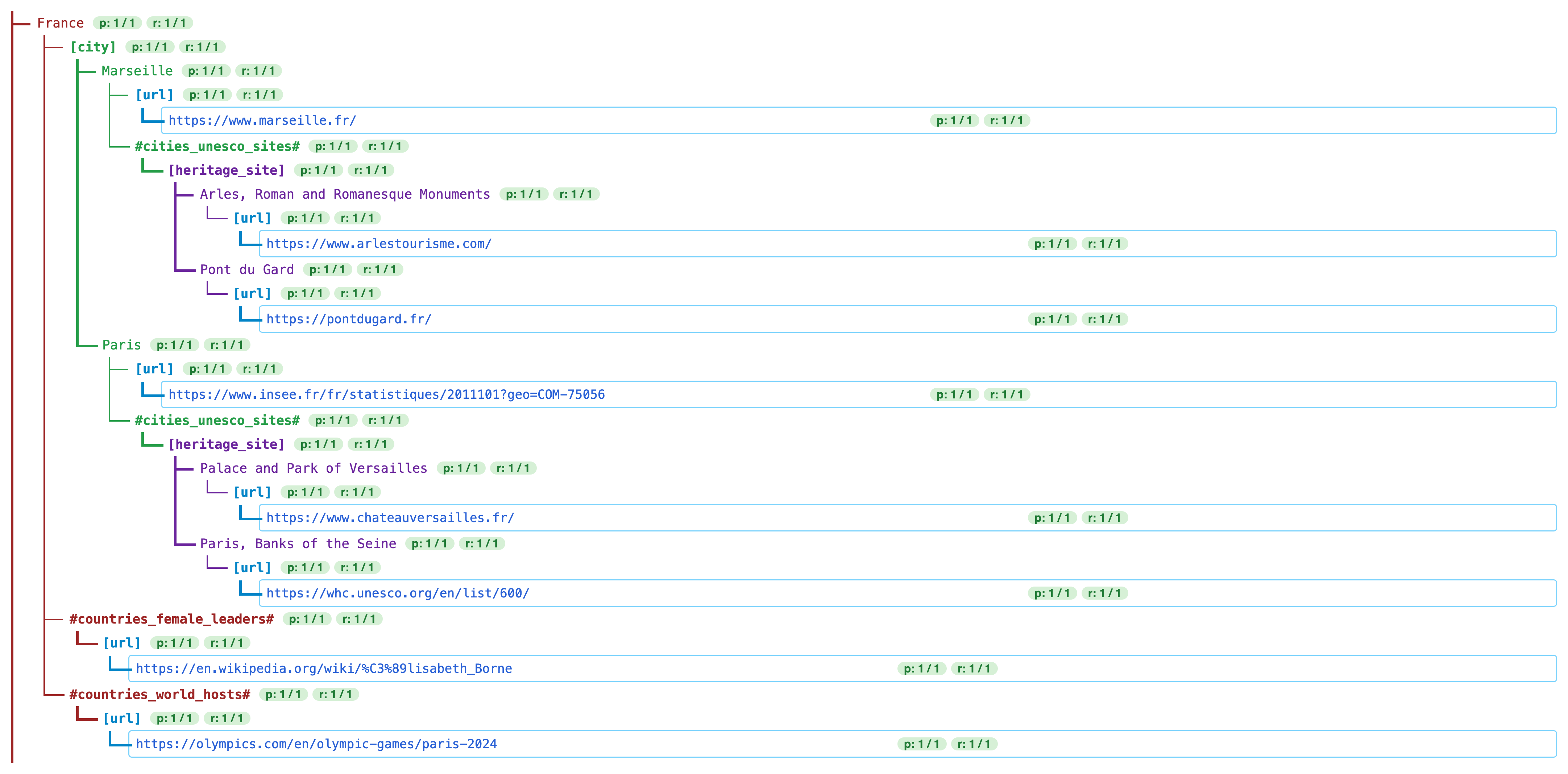}
    \caption{France is fully complete: both qualifying cities have two heritage sites, and both country-level subtasks pass. All displayed precision and recall values equal one.}
    \label{fig:toy-france}
\end{figure}

\begin{figure}[h]
    \centering
    \includegraphics[width=\linewidth]{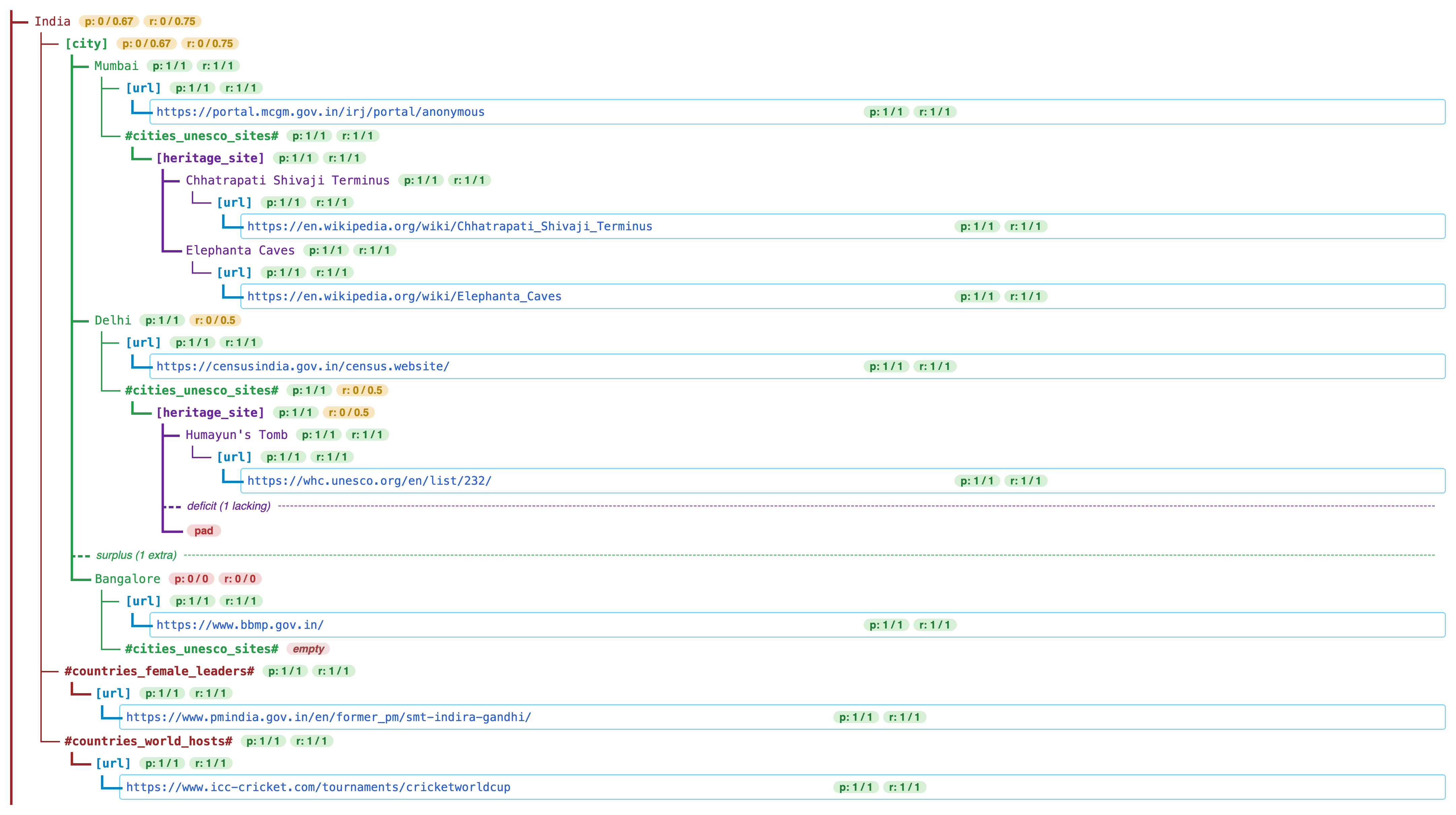}
    \caption{India illustrates partial recall and surplus handling. Delhi supplies one of two required heritage sites, so zero-padding leaves soft recall at $0.5$ while soft precision remains $1$. Three cities are supplied for a required count of two; Bangalore's empty heritage subtask gives it a zero composed score, so it is excluded when recall retains the two highest-scoring cities.}
    \label{fig:toy-india}
\end{figure}

\begin{figure}[h]
    \centering
    \includegraphics[width=\linewidth]{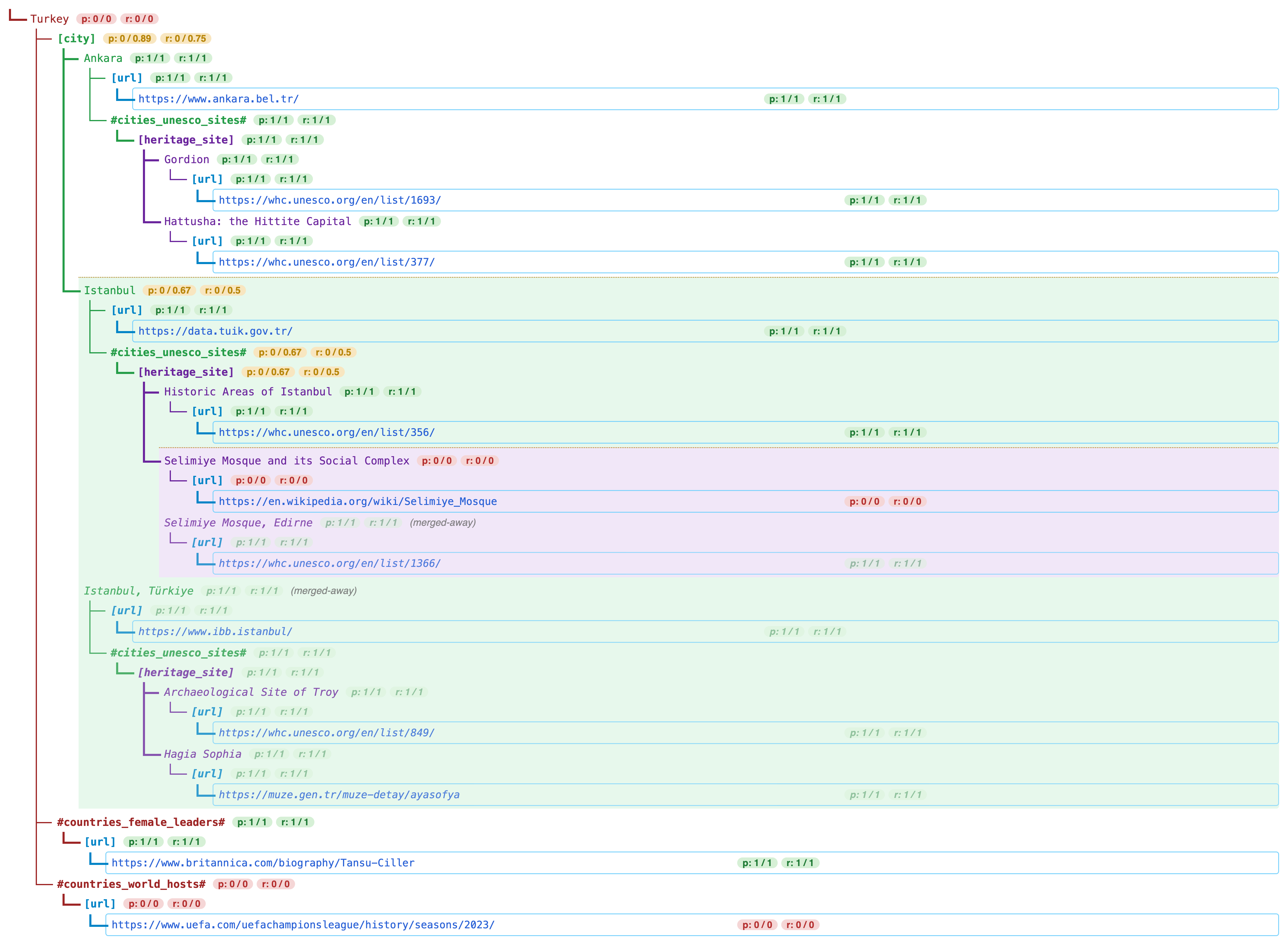}
    \caption{Turkey illustrates identity collapse and subtask gating. At the heritage level, take-worst retains an ambiguous zero-scoring citation when it merges the two Selimiye Mosque variants, reducing Istanbul's recall. The city-level aliases \texttt{Istanbul} and \texttt{Istanbul, T\"urkiye} then merge, propagating the lower score. The worldwide-sporting-event subtask also fails because its evidence concerns a continental final, so multiplicative composition sets the country score to zero.}
    \label{fig:toy-turkey}
\end{figure}

We illustrate the rollup with a toy composite task scored end to end by the released grader and rendered in the submission viewer included with each task package. The task requests three countries, two cities with more than one million inhabitants per country, and one supporting page per city---\texttt{country (3) $\rightarrow$ city (2) $\rightarrow$ url (1)}. It adds three subtasks: a per-city \texttt{heritage\_site (2) $\rightarrow$ url (1)} branch requesting two UNESCO-linked heritage sites, plus two per-country URL-only branches requesting evidence of current or historical female leadership and of hosting a worldwide sporting event. Score indicators show hard and soft precision and recall under \path{verdict_full}. The submission covers the three countries at deliberately varied levels of completeness and obtains soft F1 $0.57$ and hard F1 $0.33$ (Figure~\ref{fig:toy-overview}). Figures~\ref{fig:toy-france}--\ref{fig:toy-turkey} expand the three country subtrees.

The Turkey subtree isolates the two identity effects summarized in Table~\ref{tab:identity-collapse}: duplicate collapse reduces the distinct count, and take-worst can lower the retained cluster score before canonical identities match main-task and subtask evidence branches.

\clearpage
\subsection{Output-Delivery Ablation} \label{appendix:output-delivery}
The output ablation compares solutions delivered through sandbox files, a file-sharing tool, or output tokens. All six rows use the same 45 tasks and the same score and operational columns as the full sweep. These are independent rollouts rather than paired deterministic executions, so the differences are descriptive associations with both the delivery configuration and the particular rollout.

\begin{table}[H]
\centering
\small
\begin{tabular}{|cc|c|ccc|ccc|}
\hline
\multirow{2}{*}{\textbf{System}} & \multirow{2}{*}{\textbf{Delivery}} & \multirow{2}{*}{\textbf{Completed}} & \multicolumn{3}{c|}{\textbf{Soft}} & \multicolumn{3}{c|}{\textbf{Hard}} \\
& & & \textbf{Precision} & \textbf{Recall} & \textbf{F1} & \textbf{Precision} & \textbf{Recall} & \textbf{F1} \\
\hline
Perplexity & \path{share}   & \textbf{45} & \textbf{0.414} & \textbf{0.391} & \textbf{0.397} & \underline{0.162} & \textbf{0.160} & \textbf{0.156} \\
Perplexity & \path{output}  & \textbf{45} & 0.331 & 0.224 & 0.238 & \textbf{0.169} & \underline{0.116} & \underline{0.121} \\
\hline
Anthropic  & \path{sandbox} & \textbf{45} & \underline{0.365} & \underline{0.238} & \underline{0.262} & 0.155 & 0.095 & 0.099 \\
Anthropic  & \path{output}  & \underline{42} & 0.305 & 0.176 & 0.202 & 0.116 & 0.073 & 0.080 \\
\hline
OpenAI     & \path{sandbox} & \textbf{45} & 0.177 & 0.150 & 0.153 & 0.076 & 0.081 & 0.073 \\
OpenAI     & \path{output}  & \textbf{45} & 0.189 & 0.109 & 0.122 & 0.087 & 0.068 & 0.068 \\
\hline
\end{tabular}
\caption{Score comparison for the matched 45-task output-delivery ablation at each system's selected configuration. Bold marks the best value in each numeric column and underline marks the second-best distinct value.}
\label{tab:output-ablation-scores}
\end{table}

\begin{table}[H]
\centering
\small
\begin{tabular}{|cc|c|c|cc|cc|}
\hline
\multirow{2}{*}{\textbf{System}} & \multirow{2}{*}{\textbf{Delivery}} & \multirow{2}{*}{\textbf{Completed}} & \multicolumn{1}{c|}{\textbf{Cost}} & \multicolumn{2}{c|}{\textbf{Latency}} & \multicolumn{2}{c|}{\textbf{Token usage}} \\
& & & \textbf{\$/task} & \textbf{Med. min} & \textbf{P90 min} & \textbf{In M} & \textbf{Out k} \\
\hline
Perplexity & \path{share}   & \textbf{45} & 4.75 & 15.6 & 36.9 & 4.11 & \textbf{28.2} \\
Perplexity & \path{output}  & \textbf{45} & 5.08 & 15.4 & 33.4 & 3.79 & 48.5 \\
\hline
Anthropic  & \path{sandbox} & \textbf{45} & 49.09 & 78.5 & 132.7 & 57.04 & 316.6 \\
Anthropic  & \path{output}  & \underline{42} & 38.05 & 64.7 & 227.3 & 39.54 & 305.1 \\
\hline
OpenAI     & \path{sandbox} & \textbf{45} & \textbf{0.49} & \underline{8.8} & \underline{13.8} & \textbf{0.14} & \underline{32.1} \\
OpenAI     & \path{output}  & \textbf{45} & \underline{0.68} & \textbf{8.8} & \textbf{11.9} & \underline{0.15} & 49.5 \\
\hline
\end{tabular}
\caption{Operational statistics for the same matched output-delivery ablation as Table~\ref{tab:output-ablation-scores}. Cost and token totals are divided by 45 scheduled tasks; median and P90 are solve-stage latencies over tasks with valid timestamps. In M and Out k denote millions and thousands of tokens. Bold marks the best value and underline marks the second-best distinct value, using unrounded values; higher is better for completion and lower is better for resource use.}
\label{tab:output-ablation-operational}
\end{table}

Output-token runs have lower soft F1 than file-delivery runs for Perplexity ($0.397$ versus $0.238$), Anthropic ($0.262$ versus $0.202$), and OpenAI ($0.153$ versus $0.122$); hard F1 shows the same ordering ($0.156$ versus $0.121$, $0.099$ versus $0.080$, and $0.073$ versus $0.068$). The absolute soft F1 loss decreases monotonically with the file-delivery score across these three systems ($0.159$, $0.060$, and $0.031$). This three-system association is descriptive rather than causal, but it suggests that output-only systems may be somewhat disadvantaged while also making it unlikely that delivery alone explains the much larger gaps in the main ranking. Operational differences vary by system: output-token runs are slightly more expensive for Perplexity and OpenAI and less expensive per scheduled task for Anthropic, whose output-token run also produces fewer metric-bearing trials ($42/45$ versus $45/45$).

\clearpage
\subsection{Suitability for Reinforcement Learning} \label{appendix:rl-training}

WANDR is released primarily as an evaluation benchmark. Its structure nevertheless offers several properties relevant to future training splits and harnesses.

\paragraph{Scalable task supply and diversity} The semi-automated task pipeline avoids exhaustive answer annotation through a reusable author--critic--linter process and reference-free verification. It can produce new tasks and controlled siblings across domains, hierarchy shapes, evidence requirements, traps, and breadth/depth settings, while curation steers the resulting distribution (Section~\ref{sec:task}). This creates a path to training-scale data with broader variation than repeated sampling of the 500 released tasks. For RL-scale task supply, authoring and admission can run fully automatically, including feasibility and judge-quality gates; periodic human audits can monitor distribution drift without sitting on the generation path. To preserve evaluation validity, RL should use newly generated or separately held-out task packages, not the released benchmark split.

\paragraph{A natural curriculum} Workload is explicit in each task's key hierarchy: every level declares a required count. Holding the task's semantic criteria fixed while lowering one or more counts produces a smaller instance---for example, requiring 5 qualifying entities rather than 20, or one corroborating source rather than three---and the counts can be increased across epochs toward the full task. The same mechanism can stage breadth and depth separately, allowing a curriculum to move from reliable single-record evidence construction to sustained collection building. Count reduction controls workload rather than guaranteeing monotonic semantic difficulty, and each generated variant should still pass the normal feasibility and judge checks.

\paragraph{Dense, decomposable reward} A long rollout need not receive only a terminal all-or-nothing score. Each submitted record receives \path{verdict_retrieval} and \path{verdict_full}, and the score tree aggregates those verdicts by entity and hierarchy level into soft and hard precision, recall, and F1. Soft metrics reward correct partial progress when the requested collection is incomplete; hard metrics preserve pressure to complete whole members; and the gap between the two verdicts separates page-level task satisfaction from complete record construction. A trainer can therefore assign outcome rewards at record and branch granularity or use universal and task-specific sub-verdicts as auxiliary signals while retaining task-level F1 as the objective. This supervision is denser than a single exact-match reward, though moving from outcome-level to process supervision would require nontrivial trajectory backtracking and heuristics for assigning credit from individually graded final records to the actions responsible for them.

\paragraph{Amortized batch grading} Reference-free verification avoids constructing and maintaining exhaustive gold collections. At runtime, expensive grading remains record-local until canonicalization, deduplication, and score rollup. The released grader accepts multiple task roots, flattens their records into one streaming queue graph, and executes fetch, triage, canonicalization, deduplication, and judgment with stage-specific concurrency. Fetches are batched, and queue nodes use both in-memory request coalescing and persistent caches keyed to the smallest reusable unit. Records citing the same URL can share one fetch result; identical triage, canonicalization, and judgment work can also be reused when task semantics and inputs match. This structure aligns with RL batches, where multiple tasks and rollouts may revisit the same pages and entities. Overlap among rollout records can therefore amortize grading work, while persistent caches limit resumed runs to missing work.

\begin{table}[!htbp]
\centering
\scriptsize
\renewcommand{\arraystretch}{1.0}
\begin{tabular}{@{}>{\raggedright\arraybackslash}p{0.24\linewidth}>{\raggedright\arraybackslash}p{0.42\linewidth}>{\raggedright\arraybackslash}p{0.24\linewidth}@{}}
\toprule
\textbf{Training need} & \textbf{WANDR affordance} & \textbf{Caveat / required control} \\
\midrule
Held-out task supply & The task pipeline can generate siblings across domains, hierarchy shapes, source classes, traps, and breadth/depth settings without gold-answer enumeration. & Training should use generated or separately held-out packages, not the 500-task benchmark split. \\
Curriculum & Per-level \texttt{required} counts can be lowered or raised to stage breadth and depth separately. & Count reduction changes workload, not necessarily semantic difficulty; variants still need feasibility and judge checks. \\
Dense reward & \path{verdict_retrieval}, \path{verdict_full}, page/excerpt sub-verdicts, tree-level soft scores, and hard completion provide branch- and record-level outcome rewards. & Process supervision requires trajectory backtracking and action-level credit-assignment heuristics; reward shaping should be checked against task-level F1. \\
Batch grading & The grader's record stream, staged concurrency, URL fetch sharing, and persistent caches can amortize overlapping records across tasks and rollouts. & The public entry point should add rollout/case identifiers and grouped score reporting for native task-by-rollout batches. \\
Reward-hacking audit & The verifier exposes over-submission, duplicate clusters, excerpt failures, page failures, and source-role failures. & Training experiments should explicitly test templated evidence, judge exploitation, cache artifacts, and benchmark-split leakage. \\
\bottomrule
\end{tabular}
\caption{RL-oriented view of WANDR's task and grader design.}
\label{tab:rl-training}
\end{table}

\end{document}